\documentclass{article}
\usepackage{arxiv}

\usepackage[utf8]{inputenc} 
\usepackage[T1]{fontenc}    
\usepackage{hyperref}       
\usepackage{url}            
\usepackage{booktabs}       
\usepackage{amsfonts}       
\usepackage{nicefrac}       
\usepackage{microtype}      
\usepackage{graphicx}
\usepackage{natbib}
\usepackage{doi}
\usepackage{caption}
\usepackage{graphicx}
\usepackage{amsmath,amssymb,amsfonts, bm}
\usepackage{algorithmic}
\usepackage{algorithm}

\title{Robust Federated Learning with Global Sensitivity Estimation for Financial Risk Management}

\author{ {Lei Zhao}\\
	Department of Electrical and Computer Engineering\\
	University of Victoria\\
	\texttt{leizhao@uvic.ca} \\
	\And
	{Lin Cai} \\
	Department of Electrical and Computer Engineering\\
	University of Victoria\\
	\texttt{cai@ece.uvic.ca} \\
	\And
	{Wu-Sheng Lu} \\
Department of Electrical and Computer Engineering\\
University of Victoria\\
\texttt{wslu@ece.uvic.ca} \\
}

\date{}

\begin{document}
\maketitle

\begin{abstract}
	In decentralized financial systems, robust and efficient Federated Learning (FL) is promising to handle diverse client environments and ensure resilience to systemic risks. We propose Federated Risk-Aware Learning with Central Sensitivity Estimation (FRAL-CSE), an innovative FL framework designed to enhance scalability, stability, and robustness in collaborative financial decision-making. The framework's core innovation lies in a central acceleration mechanism, guided by a quadratic sensitivity-based approximation of global model dynamics. By leveraging local sensitivity information derived from robust risk measurements, FRAL-CSE performs a curvature-informed global update that efficiently incorporates second-order information without requiring repeated local re-evaluations, thereby enhancing training efficiency and improving optimization stability. Additionally, distortion risk measures are embedded into the training objectives to capture tail risks and ensure robustness against extreme scenarios. Extensive experiments validate the effectiveness of FRAL-CSE in accelerating convergence and improving resilience across heterogeneous datasets compared to state-of-the-art baselines.
\end{abstract}


\section{Introduction}
Global financial systems are inherently complex and dynamic, presenting significant challenges for achieving robust and efficient decision-making~\cite{banegas2020reserve, du2018deviations}. Institutions such as banks, pension funds, and insurance companies operate within highly volatile and interconnected environments, requiring advanced analytical tools to navigate uncertainties effectively. Building scalable and resilient models is not only critical for optimizing financial operations but also essential for maintaining systemic stability and ensuring long-term financial sustainability.

Traditional centralized financial modeling approaches often struggle to generalize across diverse market conditions due to data fragmentation, regulatory constraints, and institution-specific risk profiles. The heterogeneity in financial decision-making processes further complicates the development of unified predictive models, as institutions rely on distinct data sources and proprietary methodologies. Additionally, extreme market fluctuations introduce systemic risks that conventional optimization frameworks fail to address effectively, leading to potential instability in financial decision-making~\cite{lee2020learning, gagne2021two}.

Federated Learning (FL) has emerged as a promising paradigm for collaborative model training in decentralized systems~\cite{karimireddy2020scaffold, wang2020federated}. By enabling multiple institutions to train global models without sharing sensitive data, FL aligns well with the privacy and regulatory requirements of financial systems~\cite{zhao2025federated}. However, traditional FL frameworks face critical challenges in achieving robustness and efficiency. Data heterogeneity among institutions, arising from diverse market conditions and operational priorities, can cause model divergence during local training~\cite{zhao2024tailored}. Additionally, frequent communication between clients and the central server in decentralized settings leads to high communication overhead and low convergence speed. These challenges are further compounded when the trained models must deal with rare but severe risks during extreme financial market conditions~\cite{anderson2021arbitrage}.

To address these challenges, we propose Federated Risk-Aware Learning with Central Sensitivity Estimation (FRAL-CSE), a novel FL framework designed to enhance robustness, scalability, and efficiency in decentralized financial decision-making. By integrating distortion risk measures into the training process, FRAL-CSE prioritizes critical financial risks, ensuring stability under extreme market conditions while maintaining flexibility across diverse institutions. A key innovation of FRAL-CSE is its risk-aware sensitivity estimation, which refines global model updates through a sensitivity-based central acceleration mechanism. This approach reduces reliance on frequent local updates while capturing global training dynamics, mitigating data heterogeneity, and improving model robustness. Unlike traditional FL methods that rely solely on local gradients, FRAL-CSE leverages second-order sensitivity information to optimize convergence and stability. By uniting risk-aware optimization with efficient learning dynamics, FRAL-CSE accelerates convergence and enhances resilience, making it well-suited for real-world financial applications where both scalability and risk mitigation are critical. 

The rest of this paper is organized as follows. Section~\ref{related} illustrate the related work. Section~\ref{problem} introduces the embedding of distortion risk measures into FL training objective. Section~\ref{train} presents the design of robust and efficient central acceleration procedure. Section~\ref{exp} evaluates the proposed approach's performance on financial decision-making tasks with real-market data. Section~\ref{con} concludes our work and outlines directions for future research.

\section{Related Work}
\label{related}
The challenges of achieving robust and efficient model training in decentralized financial environments~\cite{isichenko2021quantitative, karatzas2021portfolio} have driven advancements in financial risk management and sensitivity analysis~\cite{shang2021capital}. FL has emerged as a powerful approach for decentralized model training, allowing multiple clients to collaborate while preserving data privacy. However, these methods often struggle with model divergence in financial applications, where client datasets exhibit significant variability due to diverse market conditions and operational priorities. Statistical heterogeneity remains a key challenge, as differences in local data distributions introduce non-IID effects that lead to training drift and performance degradation~\cite{zhao2018federated}.

To address these limitations, several approaches have been developed. Convergence analyses~\cite{ khaled2020tighter, yang2021achieving} and bounded gradient techniques~\cite{wang2019adaptive} provide theoretical insights into stabilizing training under heterogeneous conditions. FedProx~\cite{li2020federated} mitigates data heterogeneity by incorporating a proximal term in local objectives, while SCAFFOLD~\cite{karimireddy2020scaffold} employs variance reduction techniques to correct client drift. Communication efficiency has also been extensively studied, with dynamic client sampling strategies proposed to reduce overhead~\cite{gorbunov2021marina, yang2021achieving}. However, existing methods often overlook the impact of random client availability, which introduces variance, destabilizes convergence, and limits performance in real-world FL scenarios.

Sensitivity analysis has long been a fundamental tool for evaluating model robustness by quantifying how small perturbations in parameters influence outputs~\cite{borgonovo2016sensitivity, fissler2023sensitivity}. While gradients are widely used in deep learning for optimization, their role in robustness assessment and federated interpretability has been less explored~\cite{pesenti2024differential}. Recent advancements in differential sensitivity measures extend traditional gradient-based methods by capturing the evolution of training dynamics, providing deeper insights into stability, generalization, and risk exposure in model behavior~\cite{pesenti2021cascade}. These measures have shown promise in mitigating performance degradation under distribution shifts by enabling adaptive adjustments during optimization. However, existing FL frameworks rarely incorporate sensitivity-aware strategies, limiting their ability to handle heterogeneous client distributions effectively. Our work builds on these developments by integrating differential sensitivity estimation into federated learning, addressing both robustness and scalability challenges in decentralized financial systems.

Recent research in FL has focused on addressing challenges such as model heterogeneity, representation degeneration, and personalization. For instance, FedPAC~\cite{xu2023personalized} enhances feature alignment through a shared representation and personalized classifier heads but faces limitations due to its computational overhead and reliance on stable client participation. FedDBE~\cite{zhang2024eliminating} tackles domain discrepancies by employing a Domain Bias Eliminator to improve generalization and personalization, although it struggles with scalability under resource-constrained environments. Similarly, FedGH~\cite{yi2023fedgh} provides a communication-efficient approach to handle model heterogeneity by training a generalized global prediction header; however, its reliance on consistent client availability undermines its robustness in dynamic conditions. While these methods demonstrate effectiveness in specific contexts, they often fall short in addressing the combined demands of robustness, scalability, and data heterogeneity inherent in large-scale decentralized systems, which are critical for financial decision-making.

Our work addresses these limitations by introducing a novel FL framework that integrates distortion risk measures with differential sensitivity analysis. Unlike traditional FL methods that rely on local updates and suffer from client drift under heterogeneous data distributions, FRAL-CSE introduces a sensitivity-aware optimization strategy. By leveraging aggregated sensitivity measures, our approach enables more precise global model updates, mitigating local inconsistency while preserving decentralized autonomy. This design improves the stability of federated training without compromising adaptability, allowing FRAL-CSE to scale efficiently in dynamic financial environments.

\section{Risk-Aware Optimization in Federated Learning}
\label{problem}
In the FL system, clients exchange local model parameters with a central  server, which aggregates them to a global model and distributes it back to clients to assist their financial decision making. For each client $k$,  its local data set is defined as $\mathcal{D}_k =  \{(\bm{x}_i^k, y_i^k)\}_{i=1}^{n_k}$, and the risk for an individual sample $(\bm{x}_i^k, y_i^k)$ is defined as
\begin{equation}
	R(\bm{w}; \bm{x}_i^k, y_i^k) = -y_i^k f(\bm{w}, \bm{x}_i^k),
\end{equation}
where $y_i^k \in \{-1, 1\}$ is the actionable label, and $f(\bm{w}, \bm{x}_i^k)$ gives the model's prediction for input $\bm{x}_i^k$. The sign of $R(\bm{w}; \bm{x}_i^k, y_i^k)$ determines whether or not the prediction aligns with the guidance, serving as a quantitive measure of loss of the client $k$. The global model parameters $\bm{w}$ are optimized using contributions from all clients, incorporating their specific risk profiles and local conditions.

The local loss distribution of client $k$, denoted by the random variable $X_k$, quantifies the risks associated with the model's predictions $f(\bm{w}, \cdot)$ over the local dataset $\mathcal{D}_k = \{(\bm{x}_i^k, y_i^k)\}_{i=1}^{n_k}$. This distribution is constructed from the set of sample-wise risks
\begin{equation}
	X_k = \{R(\bm{w}; \bm{x}_i^k, y_i^k)\}_{i=1}^{n_k}.
\end{equation}
It provides a comprehensive representation of the potential losses encountered by client $k$ and serves as the foundation for local risk assessment.

To accommodate the diverse risk preferences of financial institutions, we incorporate a distortion risk measure that reshapes this distribution. By emphasizing specific regions, particularly the tails, this measure ensures that extreme risk scenarios receive greater attention. This approach enhances the robustness of the model, making it more resilient to rare but significant financial fluctuations.

\subsection{The Distortion Risk Measure in Decision Making}
The goal of the FL framework is to collaboratively train a global model that accurately captures and manages risks across all participating clients. By incorporating distortion risk measures, the framework prioritizes critical risk factors, ensuring robust decision-making under varying market conditions while preserving scalability in decentralized financial environments.

To quantify risk at the client level, we define the cumulative distribution function (CDF) of the local loss distribution $X_k$ under model parameters $\bm{w}$ as
\begin{equation}
	F_{X_k, \bm{w}}(\alpha) = P(X_k \leq \alpha),
\end{equation}
which represents the probability that the risk $X_k$ does not exceed a given threshold $\alpha$.
For a finite dataset $\mathcal{D}_k = \{(\bm{x}_i^k, y_i^k)\}_{i=1}^{n_k}$, the empirical CDF is approximated as
\begin{equation}
	F_{X_k, \bm{w}}(\alpha) = \frac{1}{n_k} \sum_{i=1}^{n_k} \mathbf{1}_{[R(\bm{w}; \bm{x}_i^k, y_i^k) \leq \alpha]},
\end{equation}
where $\mathbf{1}_{[\cdot]}$ is an indicator function that equals $1$ if the condition inside is true and $0$ otherwise.
To evaluate risk at a specified confidence level, we define $\beta$ as the threshold indicating the value below which a proportion $\beta$ of the risks are expected to fall. This threshold, known as the $\beta$-quantile of the risk distribution, is expressed as
\begin{equation}
	F_{X_k, \bm{w}}^{-1}(\beta) = \min \{\alpha \mid F_{X_k, \bm{w}}(\alpha) \geq \beta\}.
\end{equation}

The threshold $F_{X_k, \bm{w}}^{-1}(\beta)$ derived from the CDF serves as an upper bound on the risks, ensuring that
\begin{equation}
	R(\bm{w}; \bm{x}_i^k, y_i^k) \leq F_{X_k, \bm{w}}^{-1}(\beta),
\end{equation}
for the $\beta$-quantile of the distribution. This formulation emphasizes the importance of quantifying and controlling risks within a specified confidence level, providing a rigorous foundation for robust financial decision-making.
While this formulation ensures that risks remain below $F_{X_k, \bm{w}}^{-1}(\beta)$, it is insufficient for managing risks that exceed the threshold.

The distortion risk measure prioritizes risks in the tail of the distribution, focusing on the worst-case scenarios, which is formulated as
\begin{equation}
	\rho_k(X_k, \bm{w}) = \frac{1}{1-\beta} \int_{F_{X_k, \bm{w}}^{-1}(\beta)}^\infty R(\bm{w}; \bm{x}_i^k, y_i^k) \, dF_{X_k, \bm{w}}(\alpha).
\end{equation}
Since the full distribution $X_k$ is not directly accessible, client $k$ approximates the distortion risk measure empirically using  its local dataset $\mathcal{D}_k = \{(\bm{x}_i^k, y_i^k)\}_{i=1}^{n_k}$. The empirical approximation involves identifying the samples where risks exceed $F_{X_k, \bm{w}}^{-1}(\beta)$ which is defined as
\begin{equation}
	\mathcal{I}_\beta = \{ i \mid R(\bm{w}; \bm{x}_i^k, y_i^k) > F_{X_k, \bm{w}}^{-1}(\beta) \}.
\end{equation}
The distortion risk measure is then computed as the average of the tail risks defined as
\begin{equation}
	\rho_k(X_k, \bm{w}) \approx \frac{1}{|\mathcal{I}_\beta|} \sum_{i \in \mathcal{I}_\beta} R(\bm{w}; \bm{x}_i^k, y_i^k),
\end{equation}
which ensures that the model effectively prioritizes the most critical risks, providing a robust mechanism for decision-making under uncertainty.  Emphasizing extreme scenarios, the distortion risk measure guides the training and optimization of robust financial models.

\subsection{Robust Objective Formulation for Federated Optimization}
To effectively manage risks exceeding the upper bound $F_{X_k, \bm{w}}^{-1}(\beta)$ and enhance robustness, we introduce nonnegative slack variables $\bm{z}^k = \{z_1^k, \cdots, z_{n_k}^k\}$. These slack variables allow for a relaxation of the upper bound constraint on risks, capturing the extent to which the risk surpasses the threshold, such that
\begin{equation}
	R(\bm{w}; \bm{x}_i^k, y_i^k) \leq F_{X_k, \bm{w}}^{-1}(\beta)
\end{equation}
can be relaxed to 
\begin{equation}
	R(\bm{w}; \bm{x}_i^k, y_i^k) \leq F_{X_k, \bm{w}}^{-1}(\beta) + z_i^k,
\end{equation}
where $z_i^k$ quantifies the excessive risk over the threshold $F_{X_k, \bm{w}}^{-1}(\beta)$. By incorporating these excessive risks into the objective function, we adjust the optimization problem to balance the regularization term with penalization of slack variables, controlled by a trade-off parameter $c > 0$. This enables the model to manage risks beyond the threshold effectively.
For each client $k$, the local objective is then formulated as
\begin{equation}
	\begin{aligned}
		&\underset{(\bm{w}, \bm{z}^k)}{\text{minimize}} \quad \frac{1}{2} \|\bm{w}\|^2_2 + \frac{c}{n_k} \sum_{i=1}^{n_k} z_i^k \\
		&\text{subject to:} \quad z_i^k \geq R(\bm{w}; \bm{x}_i^k, y_i^k) - F_{X_k, \bm{w}}^{-1}(\beta), \\
		&\quad \quad \quad \quad \quad z_i^k \geq 0, \quad \forall (\bm{x}_i^k, y_i^k) \in \mathcal{D}_k,
	\end{aligned}
	\label{risk-2}
\end{equation}
minimizing the sum of the slack variables effectively reduces the average of the worst-case risks exceeding the threshold $F_{X_k, \bm{w}}^{-1}(\beta)$. This approach aligns  with the objective of  distortion risk measure  minimization, which focuses on managing extreme risks.

To refine the optimization for model training, we focus on minimizing the upper bound violation, which is equivalent to find the smallest slack variables $\{z_i^k\}_{i=1}^{n_k}$. Based on (\ref{risk-2}), it is equivalent to minimizing the cumulative slack variable penalty with the objective as
\begin{equation}
	\text{minimize} \quad \frac{c}{n_k} \sum_{i=1}^{n_k} z_i^k,
\end{equation}
which can be rewritten as 
\begin{equation}\nonumber
	\text{minimize} \quad \frac{c}{n_k} \sum_{i=1}^{n_k} \max\{0, R(\bm{w}; \bm{x}_i^k, y_i^k) - F_{X_k, \bm{w}}^{-1}(\beta)\}.
\end{equation}
Slack variables are zero for samples where $R(\bm{w}; \bm{x}_i^k, y_i^k) \leq F_{X_k, \bm{w}}^{-1}(\beta)$, contributing nothing to the training objective. For samples where $R(\bm{w}; \bm{x}_i^k, y_i^k) > F_{X_k, \bm{w}}^{-1}(\beta)$, the term $R(\bm{w}; \bm{x}_i^k, y_i^k) - F_{X_k, \bm{w}}^{-1}(\beta)$ captures the excessive risk. Thus, the  distortion risk measure  objective can be approximated as
\begin{equation}
	\rho_k(X_k, \bm{w}) \approx \frac{1}{n_k} \sum_{i=1}^{n_k} \max(0, R(\bm{w}; \bm{x}_i^k, y_i^k) - F_{X_k, \bm{w}}^{-1}(\beta)).
\end{equation}
The resulting local objective function for client $k$ is then defined as
\begin{equation}
	\begin{aligned}
		\underset{\bm{w}}{\text{minimize}} \quad  \mathcal{L}_k(\bm{w}) = \frac{1}{2} \|\bm{w}\|^2_2 + \frac{c}{n_k} \sum_{i=1}^{n_k} \max\{0, R(\bm{w}; \bm{x}_i^k, y_i^k) - F_{X_k, \bm{w}}^{-1}(\beta)\}.
	\end{aligned}
	\label{risk-4}
\end{equation}

The global objective function aggregates these local objectives, which is defined as
\begin{equation}
	\begin{aligned}
		\underset{\bm{w}}{\text{minimize}} \quad \mathcal{L}(\bm{w}) = 
		\sum_{k=1}^{K} \frac{n_k}{|\mathcal{D}|} \Bigg( 
		\frac{1}{2} \|\bm{w}\|^2_2 + \frac{c}{n_k} \sum_{i=1}^{n_k} 
		\max\{0, R(\bm{w}; \bm{x}_i^k, y_i^k) 
		- F_{X_k, \bm{w}}^{-1}(\beta)\} 
		\Bigg),
	\end{aligned}
	\label{risk-FL}
\end{equation}
where $|\mathcal{D}| = \sum_{k=1}^K n_k$ represents the total number of samples across all clients.

\section{Sensitivity-Based Global Optimization for Robust Central Acceleration}
\label{train}

To achieve robust and efficient training in FL, we integrate model differential sensitivity measures into the optimization process, which offering valuable insights into the training dynamics. By leveraging sensitivity information, the framework enhances the convergence and stability of the global model while ensuring that it generalizes effectively across diverse client environments.

\subsection{Risk-Aware Local Sensitivity Evaluation}
In the proposed framework, 
the local objective function $\mathcal{L}_k(\bm{w})$ is designed to balance two critical goals. First, it incorporates a regularization term to control the complexity of the model, preventing overfitting to local patterns that might not generalize across all clients. Second, it addresses high-risk scenarios by penalizing deviations from the distortion risk threshold ensuring that the model remains robust even under extreme conditions.
The regularization term stabilizes training by limiting the magnitude of model parameters. Its gradient with respect to the model parameters $\bm{w}$ is given by
\begin{equation}
	\nabla_{\bm{w}} \left( \frac{1}{2} \|\bm{w}\|^2_2 \right) = \bm{w},
\end{equation}
ensuring that the global model can adapt to varying financial environments without becoming overly sensitive to localized data variations.

The central acceleration mechanism in the FL framework relies on differential sensitivity measures to enhance the precision and efficiency of global updates.
Meanwhile, the distortion risk measure refines the local objective function by emphasizing high-risk scenarios. Risks that exceed the predefined threshold $F_{X_k, \bm{w}}^{-1}(\beta)$ are penalized, ensuring that the model prioritizes mitigating extreme financial losses and improving robustness. The gradient of the penalty term is given by
\begin{equation}
	\begin{aligned}
		\nabla_{\bm{w}} \left( \frac{c}{n_k} \max\{0, R(\bm{w}; \bm{x}_i^k, y_i^k) - F_{X_k, \bm{w}}^{-1}(\beta)\} \right) = 
		\begin{cases}
			-\frac{c}{n_k} y_i^k \nabla_{\bm{w}} f(\bm{w}, \bm{x}_i^k), & \text{if } R(\bm{w}; \bm{x}_i^k, y_i^k) - F_{X_k, \bm{w}}^{-1}(\beta) > 0, \\
			0, & \text{otherwise}.
		\end{cases}
	\end{aligned}
\end{equation}
Consequently, the local model sensitivity is expressed as
\begin{equation}
	\begin{aligned}
		\nabla_{\bm{w}} \mathcal{L}_k(\bm{w}) = \bm{w} 
		- \frac{c}{n_k} \sum_{i=1}^{n_k} \mathbf{1}_{\left\{R(\bm{w}; \bm{x}_i^k, y_i^k) - F_{X_k, \bm{w}}^{-1}(\beta) > 0\right\}} \times y_i^k \nabla_{\bm{w}} f(\bm{w}, \bm{x}_i^k),
	\end{aligned}
\end{equation}
where the indicator function $\mathbf{1}_{\{\cdot\}}$ ensures that only samples with risks exceeding the threshold contribute to the gradient.

Each client customizes its local gradient updates based on its unique market dynamics and risk profile. These localized updates are then aggregated to refine the global model, 
\begin{equation}
	\begin{aligned}
		\nabla_{\bm{w}} \mathcal{L}(\bm{w}) = \frac{1}{K} \sum_{k=1}^{K} \frac{n_k}{|\mathcal{D}|} \nabla_{\bm{w}} \mathcal{L}_k(\bm{w}),
	\end{aligned}
\end{equation}
which allows the global model to be resilient to high-risk scenarios, ensuring it generalizes effectively across diverse institutional environments.

\subsection{Central Acceleration via Sensitivity-Guided Optimization}

The global model risk-aware sensitivity estimation enhances efficiency in the FL framework by prioritizing critical risks that exceed a defined threshold during central acceleration. By focusing on worst-case scenarios, this approach accelerates convergence and improves stability, ensuring robust decentralized decision-making. The client $k$ contributes localized insights from its dataset $\mathcal{D}_k$, which the central server aggregates to estimate the global risk-aware sensitivity. This estimation captures the curvature of the global objective landscape, enabling precise updates that enhance optimization efficiency across the federated network while ensuring robustness to dynamic market conditions.

For the local objective function $\mathcal{L}_k(\bm{w})$, the second-order sensitivity combines contributions from both regularization and the tail-focused loss term, which is expressed as
\begin{equation}
	\begin{aligned}
		\nabla^2_{\bm{w}} \mathcal{L}_k(\bm{w}) 
		= \nabla^2_{\bm{w}} \left(\frac{1}{2} \|\bm{w}\|^2_2\right)  + \frac{c}{n_k} \sum_{i=1}^{n_k} \nabla^2_{\bm{w}} \max\{0, R(\bm{w}; \bm{x}_i^k, y_i^k) - F_{X_k, \bm{w}}^{-1}(\beta)\}.
	\end{aligned}
\end{equation}
Based on the local second-order sensitivity information, our framework introduces an efficient approximation that prioritizes the most critical samples, i.e., those exceeding the distortion threshold $F_{X_k, \bm{w}}^{-1}(\beta)$. This selective focus allows the second-order sensitivity approximation to capture essential curvature information from high-risk scenarios.

The loss term $\max\{0, R(\bm{w}; \bm{x}_i^k, y_i^k) - F_{X_k, \bm{w}}^{-1}(\beta)\}$ penalizes predictions that exceed the distortion risk measure threshold $F_{X_k, \bm{w}}^{-1}(\beta)$, ensuring that the training process focuses on managing the most critical risks in the loss distribution's tail. Our design emphasizes robustness by allocating greater focus to extreme losses, which are of particular concern in financial decision-making.
To design the global model risk-aware sensitivity estimation, we transform the  loss term $\max\{0, R(\bm{w}; \bm{x}_i^k, y_i^k) - F_{X_k, \bm{w}}^{-1}(\beta)\}$  to a quadratic penalty only on the focused high risk region, i.e., 
\begin{equation}
	\begin{aligned}
		\mathbf{1}_{\{R(\bm{w}; \bm{x}_i^k, y_i^k) - F_{X_k, \bm{w}}^{-1}(\beta) > 0\}} \left(R(\bm{w}; \bm{x}_i^k, y_i^k) - F_{X_k, \bm{w}}^{-1}(\beta)\right)^2.
	\end{aligned}
\end{equation}
By focusing on the squared deviations from the threshold for samples that exceed $F_{X_k, \bm{w}}^{-1}(\beta)$, the model remains sensitive to the most significant risks.

To refine this approximation, we employ a first-order Taylor expansion of $f(\bm{w}, \bm{x}_i^k)$ around the current model parameters $\bm{w}_t$
\begin{equation}
	f(\bm{w}, \bm{x}_i^k) \approx f(\bm{w}_t, \bm{x}_i^k) + \nabla_{\bm{w}} f(\bm{w}_t, \bm{x}_i^k)^\top (\bm{w} - \bm{w}_t).
\end{equation}
Substituting this expansion into the loss approximation gives
\begin{equation}
	\frac{c}{n_k} \sum_{i=1}^{n_k}  \left( f(\bm{w}_t, \bm{x}_i^k) 
	+ \nabla_{\bm{w}} f(\bm{w}_t, \bm{x}_i^k)^\top (\bm{w} - \bm{w}_t) \right)^2.
\end{equation}
Letting $g_i^k = f(\bm{w}_t, \bm{x}_i^k) + \nabla_{\bm{w}} f(\bm{w}_t, \bm{x}_i^k)^\top (\bm{w} - \bm{w}_t)$, the gradient of the loss w.r.t. $\bm{w}$ becomes
\begin{equation}
	\nabla_{\bm{w}} \left( \frac{c}{n_k} \sum_{i=1}^{n_k}  (g_i^k)^2 \right) 
	= \frac{c}{n_k} \sum_{i=1}^{n_k} 2  g_i^k \nabla_{\bm{w}} f(\bm{w}_t, \bm{x}_i^k).
\end{equation}
Using the chain rule, the Hessian is approximated as
\begin{equation}
	\begin{aligned}
		\nabla^2_{\bm{w}} \left( \frac{c}{n_k} \sum_{i=1}^{n_k}  (g_i^k)^2 \right) 
		\approx \frac{c}{n_k} \sum_{i=1}^{n_k} 2  \nabla_{\bm{w}} f(\bm{w}_t, \bm{x}_i^k) 
		 \times \nabla_{\bm{w}} f(\bm{w}_t, \bm{x}_i^k)^\top.
	\end{aligned}
\end{equation}

During our analysis at the $t$-th round, we simplify $F_{X_k, \bm{w}_t}^{-1}(\beta)$ as a constant threshold w.r.t the current global model $\bm{w}_t$, and we define $\bm{J}_{t,k}$ as the Jacobian matrix for client $k$, where we have
\begin{equation}
	\begin{aligned}
		\bm{J}_{t,k}^\top \bm{J}_{t,k} =
		&\sum_{i=1}^{n_k} \mathbf{1}_{\left\{R(\bm{w}; \bm{x}_i^k, y_i^k)  - F_{X_k, \bm{w}}^{-1}(\beta) > 0\right\}} \\
		&\times \nabla_{\bm{w}} f(\bm{w}, \bm{x}_i^k) \nabla_{\bm{w}} f(\bm{w}, \bm{x}_i^k)^\top.
	\end{aligned}
\end{equation}
We define the aggregated global second-order sensitivity matrix as
\begin{equation}
	\bm{S}_t = \sum_{k=1}^K \frac{n_k}{n} \left( I + \frac{c}{n_k} \bm{J}_{t,k}^\top \bm{J}_{t,k} \right),
\end{equation}
where $\bm{J}_{t,k}$ is the Jacobian matrix of the local model at client $k$, $ n = \sum_{k=1}^K n_k $ is the total dataset size, and $ c $ is a regularization constant. The matrix $\bm{S}_t$ is fixed during the central update, capturing second-order sensitivity information from the aggregated local models. Based on the designed $\bm{S}_t$, the central acceleration updating at round $t$ is computed as
\begin{equation}
	\bm{w}_{t+1} = \bm{w}_t - \left( \bm{S}_t + \epsilon I \right)^{-1} \nabla_{\bm{w}_t} \mathcal{L}(\bm{w}),
\end{equation}
where $\nabla_{\bm{w}_t} \mathcal{L}(\bm{w})$ is the gradient of the global loss function w.r.t. the current global model $ \bm{w}_t $, and $ \epsilon I $ is added for numerical stability.

The central server utilizes this second-order update to accelerate convergence while preserving model stability. Once the updated global parameters $\bm{w}_{t+1}$ are obtained, they are distributed to clients for the next round of local training. During this phase, each client updates its local model and computes new sensitivity information, which is then aggregated to refine the global sensitivity matrix for subsequent rounds, ensuring continuous adaptation and optimization.

\section{Experiments}
\label{exp}
To evaluate the effectiveness of our FRAL-CSE framework, we conduct extensive experiments using the Jane Street Market Prediction dataset. This dataset contains high-frequency financial market data, including a rich set of derived financial metrics and binary decision labels indicating favorable or unfavorable market positions. The dataset reflects real-world financial decision-making, where institutions rely on proprietary feature engineering to extract meaningful signals from market data.

\subsection{Experimental Setup}

Each sample in the dataset is represented by 130 numerical features and a binary action label. These features are not directly observed from raw market data but are derived through institution-specific processing, incorporating proprietary domain knowledge. Due to competitive and regulatory constraints, these processed features cannot be shared across financial institutions, necessitating a FL approach to enable collaborative model training while preserving data privacy.

Although all institutions share the same 130 feature dimensions, the distribution of feature values varies significantly across clients. Differences in market focus, institutional strategies, and proprietary feature engineering techniques create distinct statistical properties for identical features. For example, institutions specializing in equities may extract different market signals from the same features compared to those focused on commodities or foreign exchange. This variation leads to feature heterogeneity, where identical feature dimensions represent different financial dynamics across institutions, resulting in highly heterogeneous local datasets.

To simulate a realistic decentralized financial environment, we partition the dataset using the ExDir partitioning strategy~\cite{li2024convergence}, which effectively captures institutional specialization while ensuring realistic statistical heterogeneity. This partitioning process consists of two stages. First, clients are assigned specific financial labels corresponding to their market sector focus, reflecting real-world institutional expertise in different asset classes. Second, within each sector, data is allocated through a Dirichlet-based non-IID sampling strategy with a concentration parameter of $1.0$, ensuring that each client's dataset exhibits unique distributional characteristics while maintaining a representative sample of market variations.

We evaluate the performance of FRAL-CSE against several state-of-the-art baseline methods, including FedKD~\cite{wu2022communication}, PFL-DA~\cite{shi2023personalized}, and FedProto~\cite{tan2022fedproto}, under diverse experimental conditions. Our comparisons focus on three key aspects, i.e., scalability, robustness to data distribution shifts, and adaptability to dynamic client participation.

\subsection{Model Accuracy and Scalability}
\begin{figure}
	\centering
	\begin{tabular}{cccc} 
		\includegraphics[width = 0.45\linewidth]{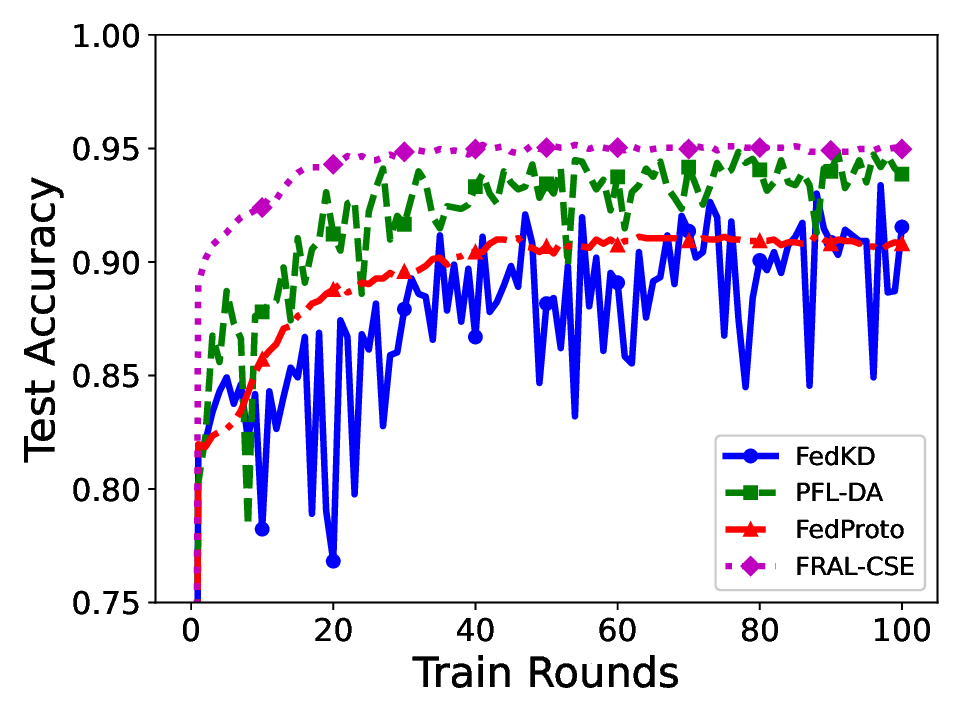}& 
		\includegraphics[width = 0.45\linewidth]{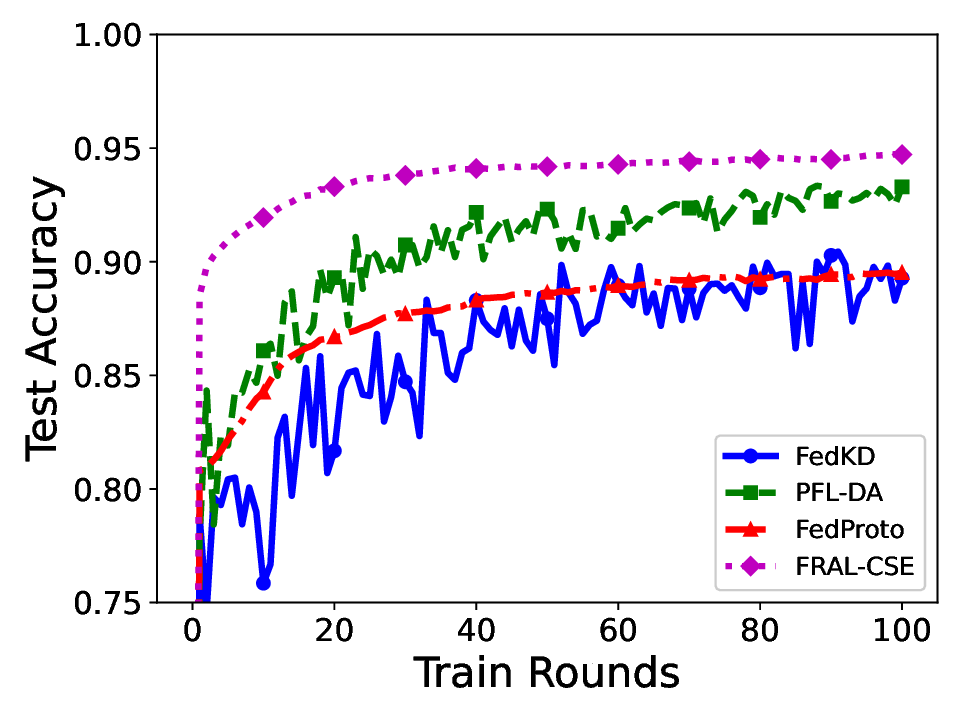} \\
		{\scriptsize (a) $10$ Clients, $10,000$ Samples Total} &
		{\scriptsize (b) $50$ Clients, $50,000$ Samples Total} \\
		\includegraphics[width = 0.45\linewidth]{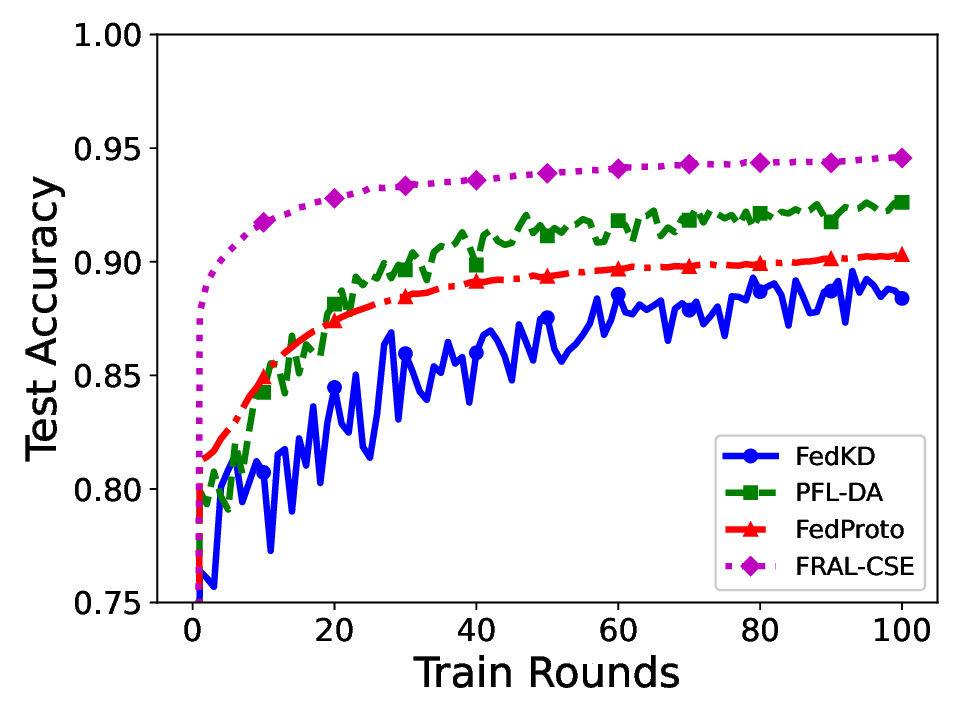} & 
		\includegraphics[width = 0.45\linewidth]{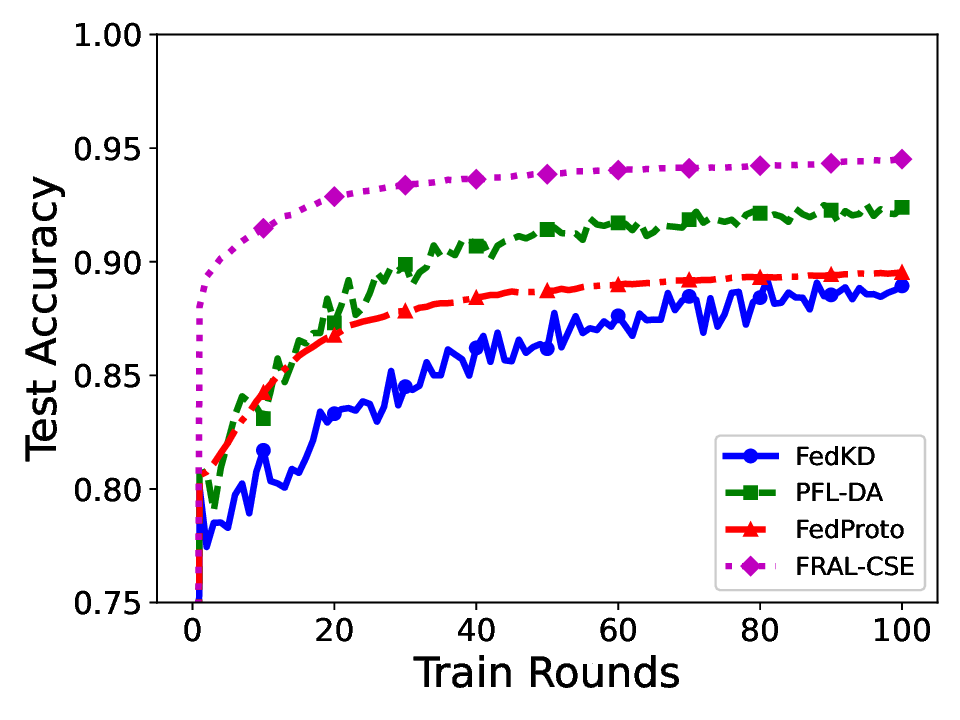}
		\\ 
		{\scriptsize (c) $100$ Clients, $100,000$ Samples Total} &
		{\scriptsize (d) $200$ Clients, $200,000$ Samples Total}  
	\end{tabular}
	\captionsetup{font={scriptsize}}
	\caption{Impact of increasing FL scales on test accuracy with full client participation and no dropout.}
	\label{fig-Scalability-acc}
\end{figure}
To assess the accuracy and scalability of FRAL-CSE, we analyze its performance as the number of participating clients increases while keeping the local dataset size fixed. This setting reflects real-world scenarios where an increasing number of financial institutions collaborate in a FL framework while maintaining data privacy. The total dataset size grows proportionally with the number of clients, ensuring that as participation expands, the collective knowledge of the system also increases. We evaluate FRAL-CSE under four different configurations, where data is distributed among 10, 50, 100, and 200 clients, with each receiving 10,000, 50,000, 100,000, and 200,000 samples, respectively.

Each client uses 80\% of its allocated data for training, while the remaining 20\% is reserved for testing. To preserve the temporal structure of financial data and create a realistic forecasting environment, we do not apply random shuffling during the train-test split. This setup ensures that test data follows training data in sequence, mirroring how financial institutions predict future market trends based on past observations.

As shown in Fig.~\ref{fig-Scalability-acc}, FRAL-CSE consistently outperforms state-of-the-art baselines across all participation scales. At the smallest scale with $10$  clients, FRAL-CSE achieves significantly higher test accuracy and converges more efficiently than all baseline methods. While the baselines exhibit slower convergence and greater variability, FRAL-CSE effectively leverages the available data to produce stable and high-performing results.

As the number of clients increases to $50$, the performance gap becomes even more pronounced. FRAL-CSE maintains consistently higher accuracy throughout the training process, whereas baseline methods, particularly FedKD and PFL-DA, struggle with slower convergence and reduced accuracy. When scaling up to 100 and 200 clients, FRAL-CSE continues to demonstrate the best test accuracy with minimal variance, effectively adapting to scaling up collaborations. While the baseline methods show slight improvements as more data becomes available, they fail to match the stability and efficiency of FRAL-CSE, particularly under conditions of high client participation.

\begin{figure}
	\centering
	\begin{tabular}{cccc} 
		\includegraphics[width = 0.45\linewidth]{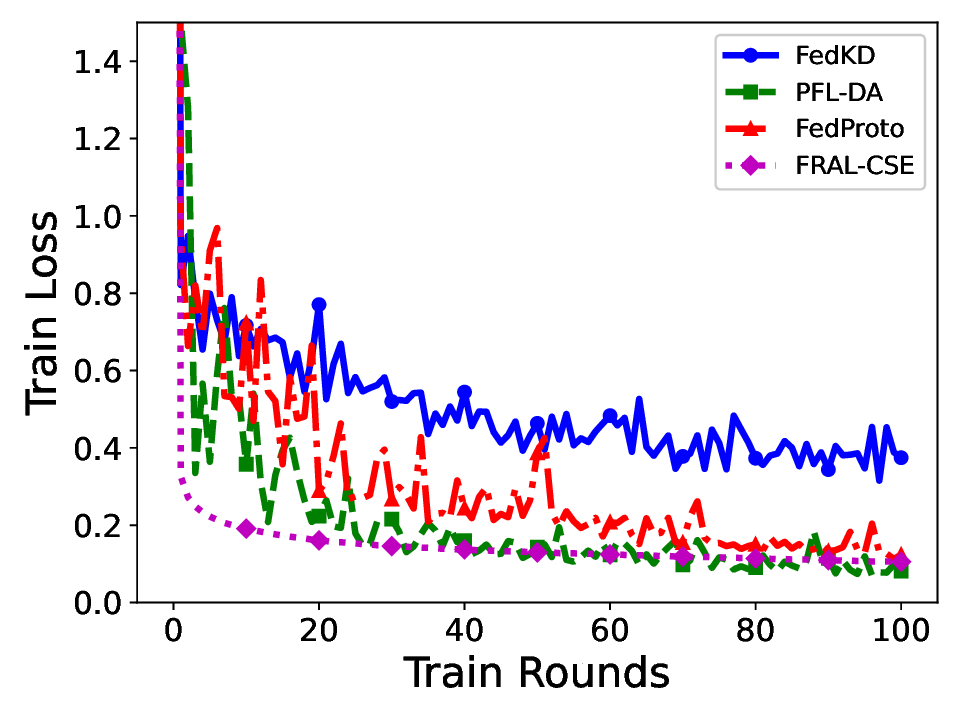}& 
		\includegraphics[width = 0.45\linewidth]{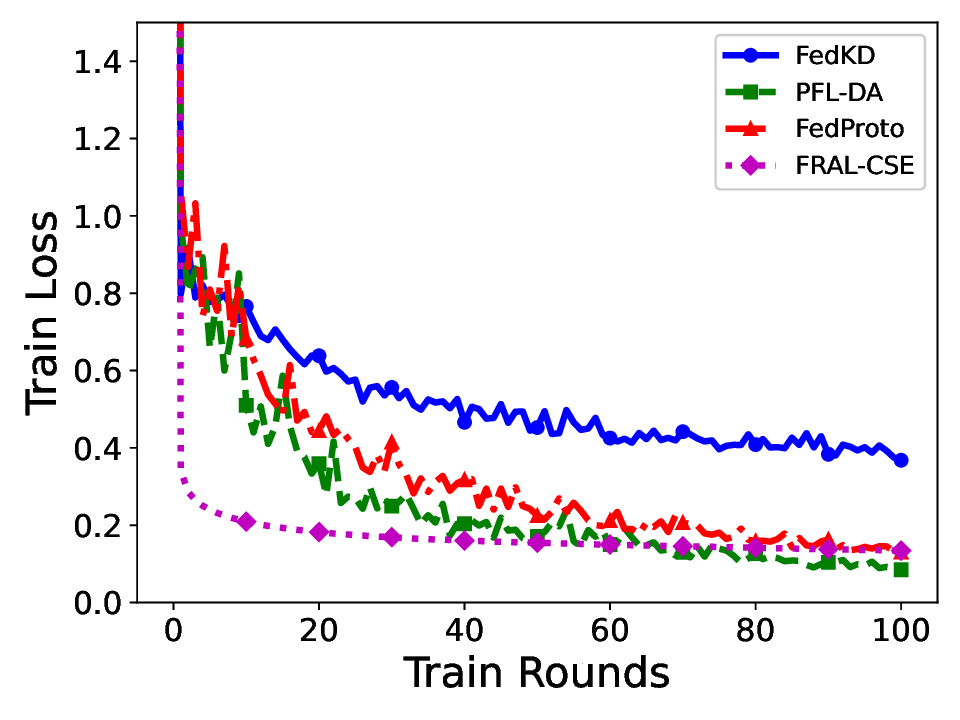} \\
		{\scriptsize (a) $10$ Clients, $10,000$ Samples Total} &
		{\scriptsize (b) $50$ Clients, $50,000$ Samples Total} \\
		\includegraphics[width = 0.45\linewidth]{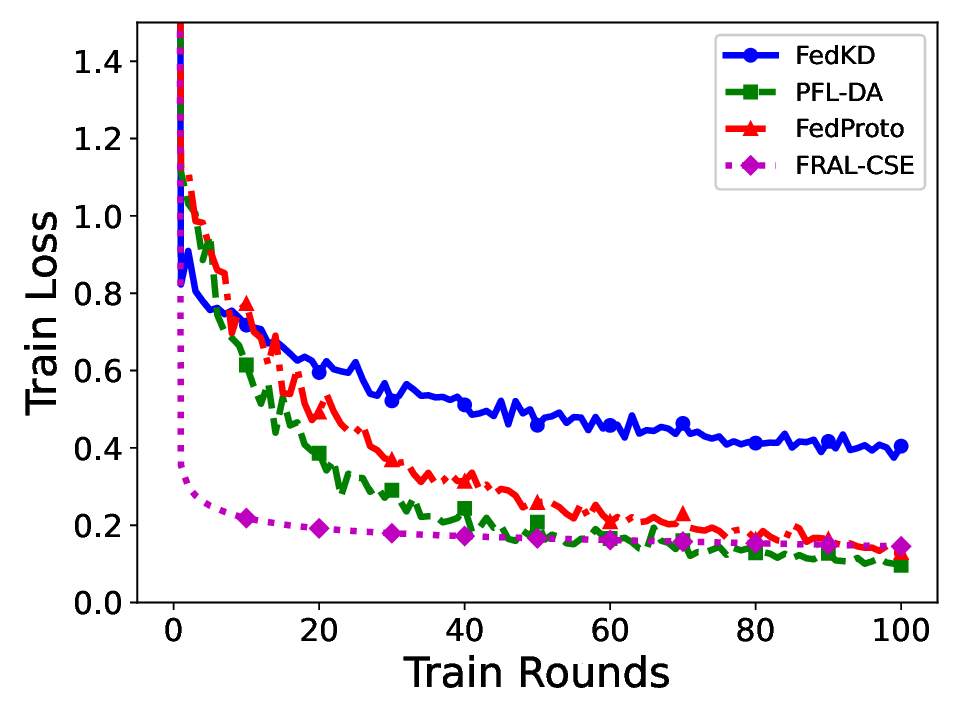} & 
		\includegraphics[width = 0.45\linewidth]{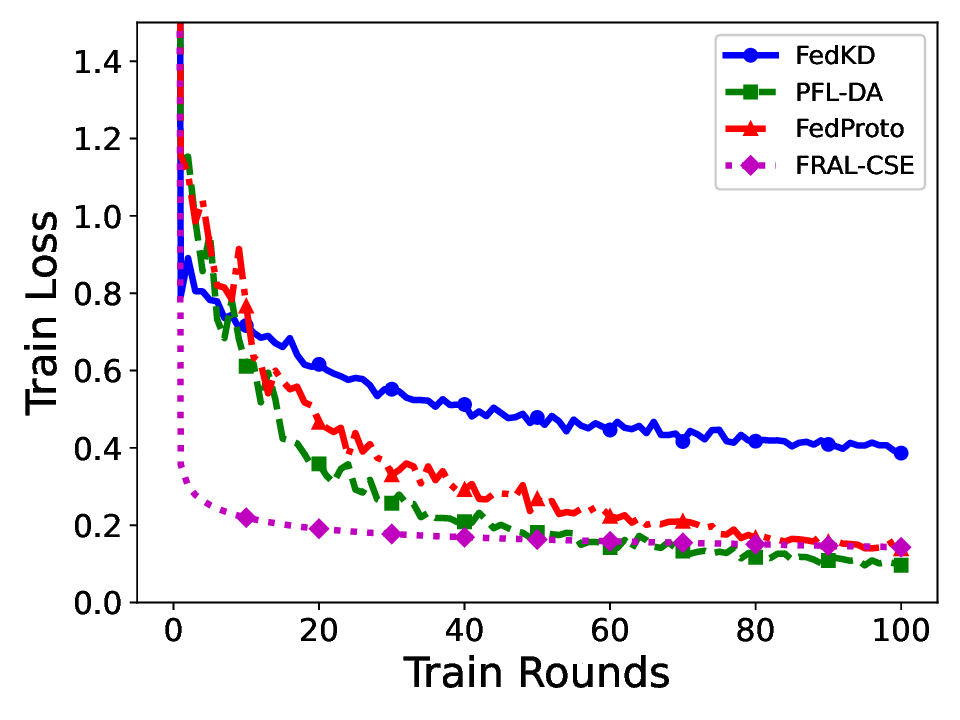} 
		\\ 
		{\scriptsize (c) $100$ Clients, $100,000$ Samples Total} &
		{\scriptsize (d) $200$ Clients, $200,000$ Samples Total}  
	\end{tabular}
	\captionsetup{font={scriptsize}}
	\caption{Impact of increasing FL scales on train loss  with full client participation and no dropout.}
	\label{fig-Scalability-loss}
\end{figure}

To further evaluate scalability, Fig.~\ref{fig-Scalability-loss} presents the corresponding training loss performance, complementing the test accuracy results in Fig.~\ref{fig-Scalability-acc}. At the 10-client scale, FRAL-CSE exhibits rapid convergence with significantly lower training loss than the baselines, which display noticeable fluctuations and slower optimization. FedKD and FedProto, in particular, require extended iterations to stabilize. As the client scale increases to 50, FRAL-CSE continues to achieve faster and more stable convergence, while the baselines show slight improvements but still suffer from prolonged optimization times and higher loss variability. FedKD and PFL-DA, in particular, require additional iterations to achieve comparable loss reduction. At 100 and 200 clients, FRAL-CSE maintains its efficiency, achieving consistently lower training loss with minimal variance. While the baselines eventually converge, they require significantly more iterations to reach competitive loss levels. Notably, PFL-DA achieves a slightly lower training loss than FRAL-CSE, but this advantage does not translate into better test accuracy. As shown in Fig.~\ref{fig-Scalability-acc}, FRAL-CSE consistently outperforms all baselines in generalization, demonstrating superior robustness to data heterogeneity and evolving financial conditions.

\subsection{Robustness to Evolving Market Conditions}

To assess the robustness of FRAL-CSE, we evaluate its generalization ability across different train-test splits. This analysis examines how well the model adapts as the temporal gap between training and test data increases, simulating real-world financial scenarios where future market conditions deviate significantly from historical trends. Smaller training ratios introduce greater distribution shifts in the test phase due to the dataset's temporal structure, making the prediction task increasingly challenging.

We conduct experiments under two train-test configurations, i.e.,  a 90\% training and 10\% testing split, and a 50\% training and 50\% testing split. These configurations are evaluated with 50 clients to analyze the model's resilience under varying levels of historical data availability.

\begin{figure}
	\centering
	\begin{tabular}{cccc} 
		\includegraphics[width = 0.45\linewidth]{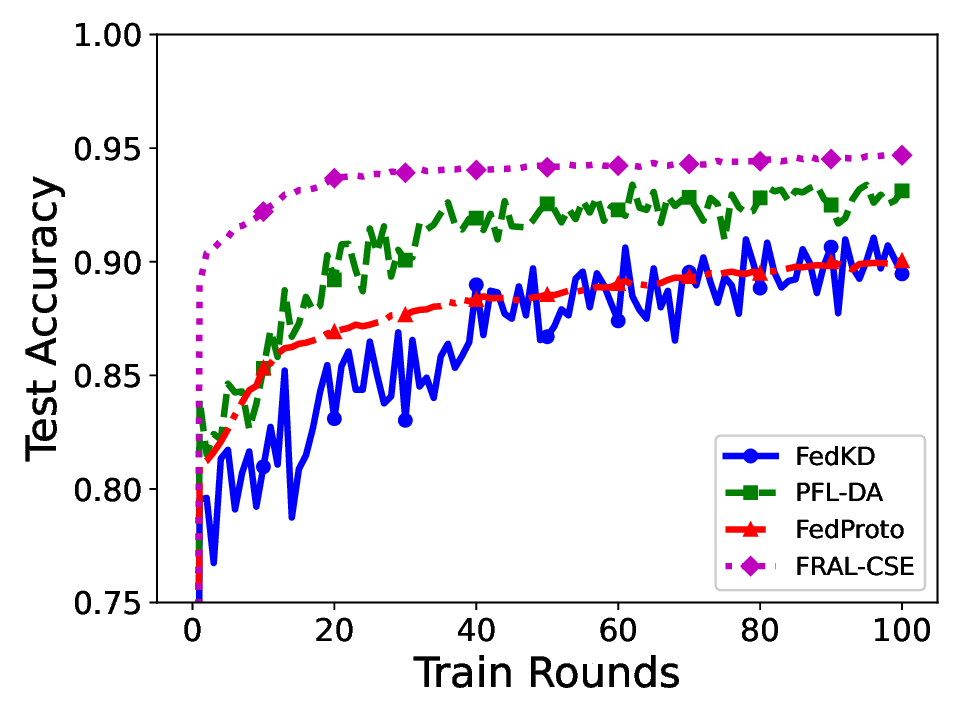} & 
		\includegraphics[width = 0.45\linewidth]{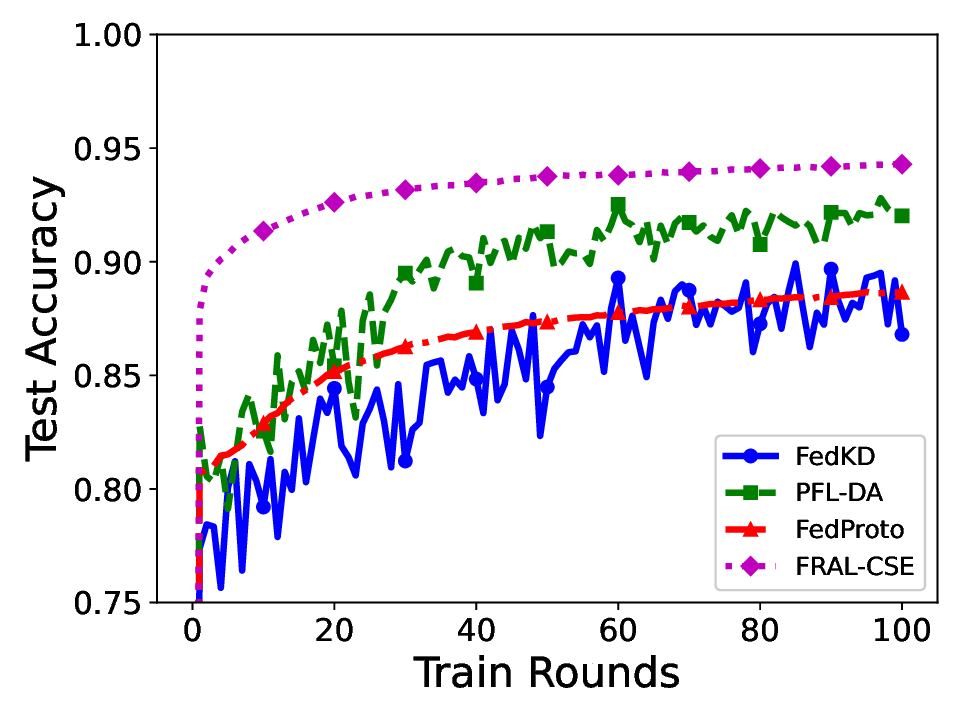} \\
		{\scriptsize (a) Training: 90\%, Testing: 10\%} &
		{\scriptsize (b) Training: 50\%, Testing: 50\%} 
	\end{tabular}
	\captionsetup{font={scriptsize}}
	\caption{Impact of different train-test splits on test accuracy with $50$ clients.}
	\label{fig-Robustness-acc-50000}
\end{figure}

As shown in Fig.~\ref{fig-Robustness-acc-50000}, FRAL-CSE consistently outperforms the baseline methods across both settings. Under the 90\% training and 10\% testing split, FRAL-CSE achieves superior accuracy with faster convergence and lower variance compared to the baselines. Even in the more challenging 50\% training and 50\% testing split, where the temporal gap between training and test data is significantly larger, FRAL-CSE maintains higher test accuracy and greater stability. In contrast, baseline methods exhibit slower convergence and increased fluctuations, underscoring their limitations in handling distribution shifts effectively.

\subsection{Robustness in Dynamic Participation and Client Dropouts}

In real-world FL applications for financial decision-making, client participation is dynamic, reflecting the operational realities of financial institutions that may join or exit the training process at different stages. This dynamic nature introduces variability in participation rates and unexpected client dropouts, both of which can affect training stability and model performance. To assess the robustness of FRAL-CSE in such scenarios, we evaluate its performance under two distinct conditions, i.e., varying participation rates and random client dropouts.

\begin{figure}
	\centering
	\begin{tabular}{cccc} 
		\includegraphics[width = 0.45\linewidth]{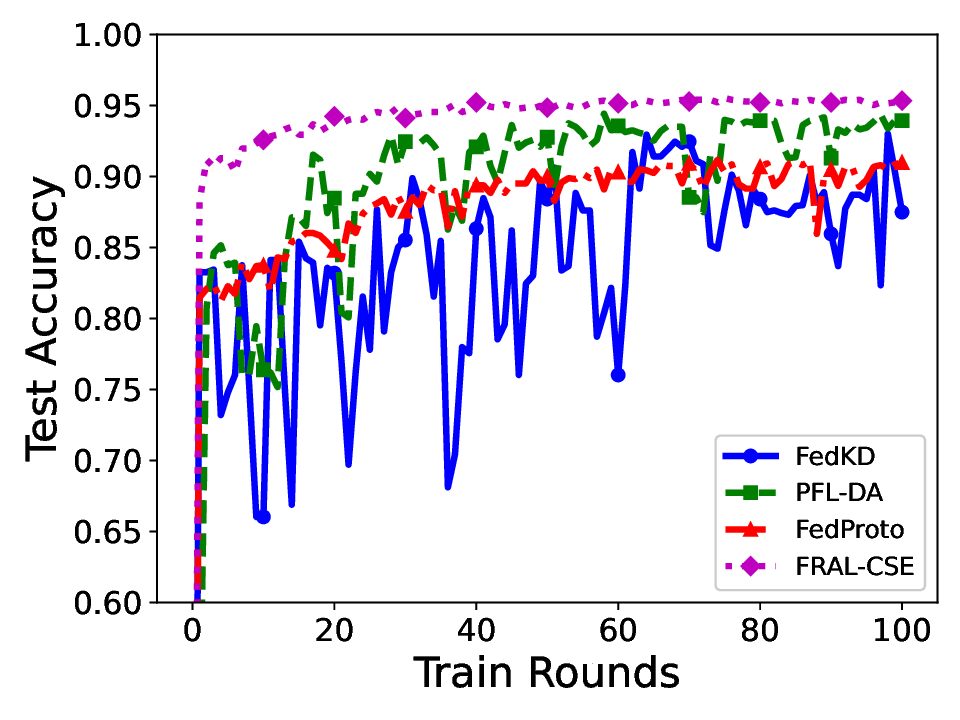} & 
		\includegraphics[width = 0.45\linewidth]{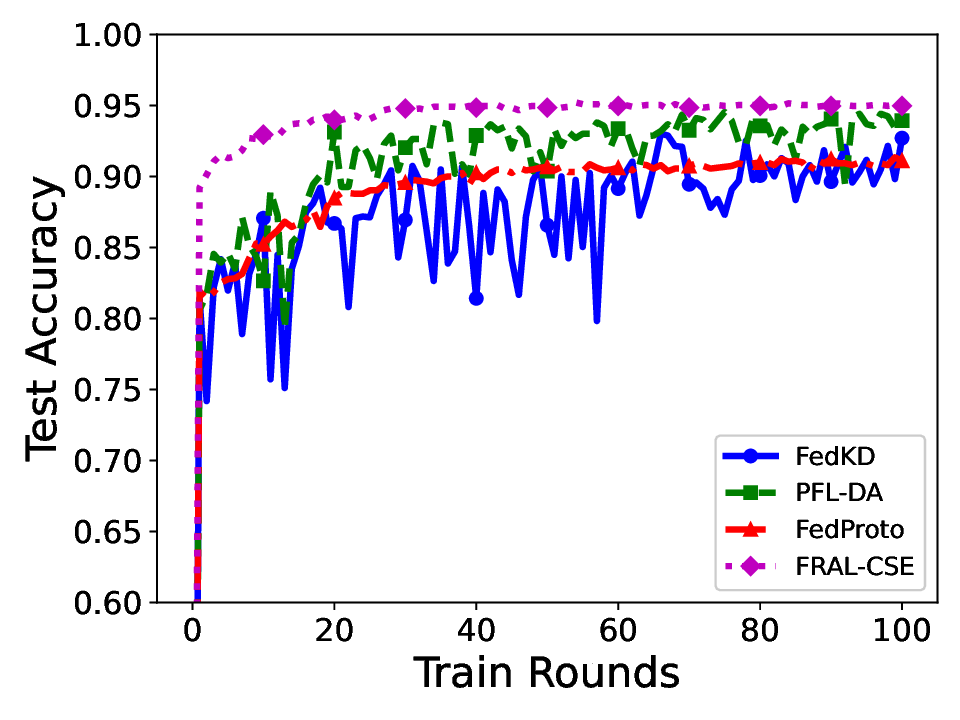} \\
		{\scriptsize (a) $20\%$ Participation Rate} &
		{\scriptsize (b) $80\%$ Participation Rate} 
	\end{tabular}
	\captionsetup{font={scriptsize}}
	\caption{Impact of different participation rates on test accuracy with 10 clients.}
	\label{fig-10000-acc-jr}
\end{figure}

First, we simulate fluctuating participation levels by fixing participation rates at 20\% and 80\%, where only a fraction of clients actively contribute to training during each communication round. As illustrated in Fig.~\ref{fig-10000-acc-jr}, which involves 10 clients in total, FRAL-CSE consistently maintains high test accuracy and stable convergence across both participation rates. Unlike baseline methods, which experience significant degradation in accuracy as fewer clients participate, FRAL-CSE demonstrates its ability to effectively utilize the available data while preserving model stability. This resilience highlights the proposed framework's adaptability to varying levels of client engagement.

\begin{figure}
	\centering
	\begin{tabular}{cccc} 
		\includegraphics[width = 0.45\linewidth]{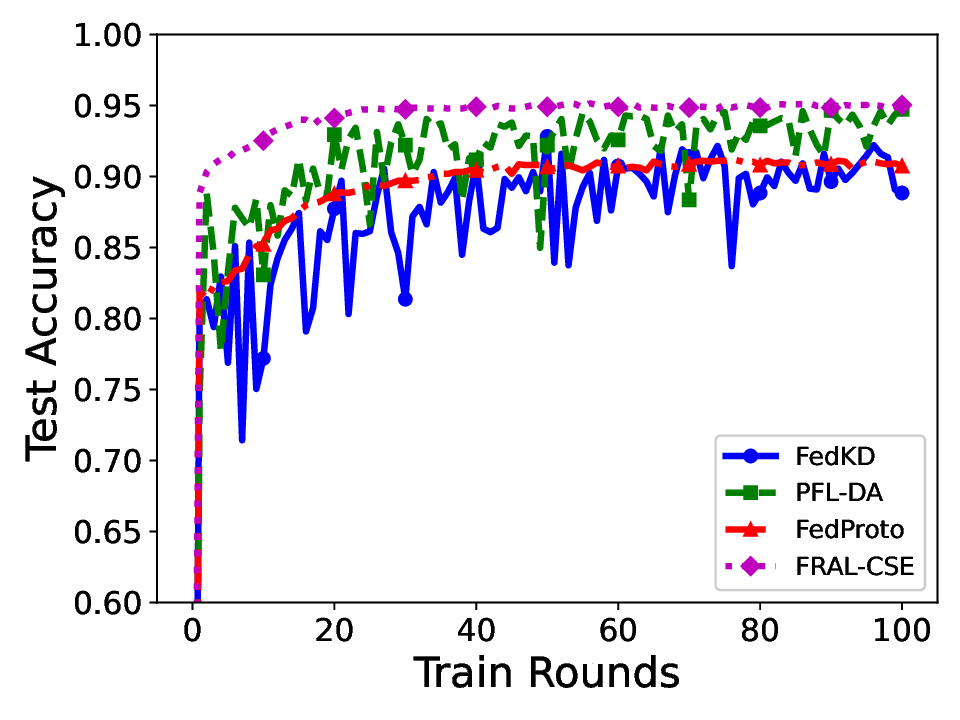} & 
		\includegraphics[width = 0.45\linewidth]{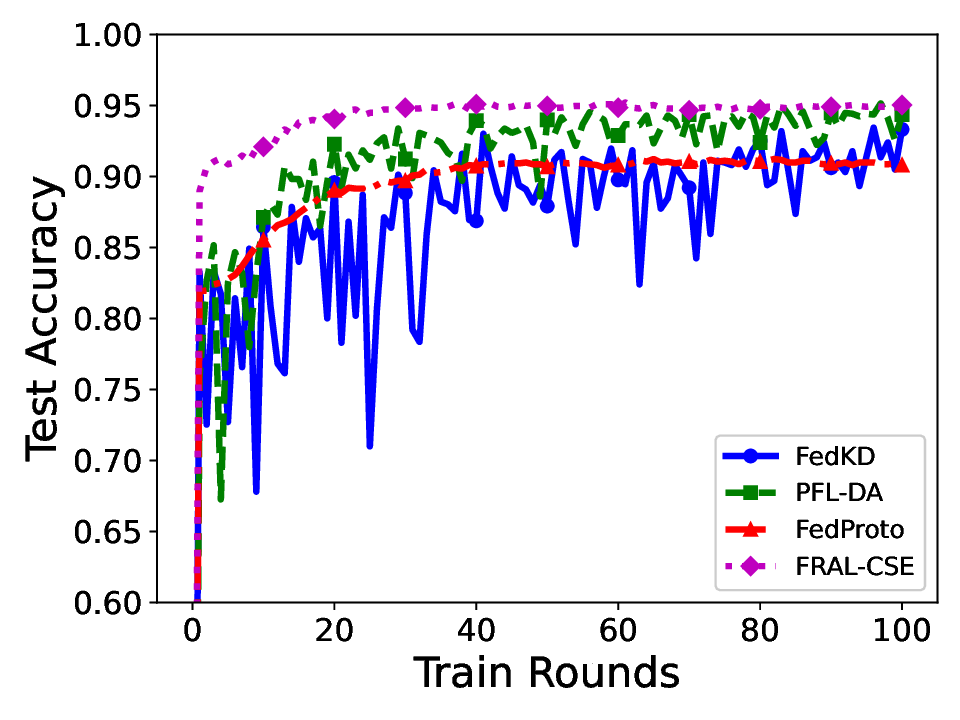} \\
		{\scriptsize (a) $10\%$ Dynamic Client Dropout} &
		{\scriptsize (b) $40\%$ Dynamic Client Dropout} 
	\end{tabular}
	\captionsetup{font={scriptsize}}
	\caption{Impact of dynamic client dropout on test accuracy with $10$ clients.}
	\label{fig-drop-out-acc-10000}
\end{figure}

Next, we simulate random client dropouts by introducing dropout rates of 10\% and 40\% during training, as shown in Fig.~\ref{fig-drop-out-acc-10000} for 10 clients. This setup closely mirrors real-world decentralized financial systems, where institutions may intermittently disengage from collaborative training due to operational or technical constraints. Despite these challenges, FRAL-CSE exhibits strong resilience, achieving stable convergence and maintaining high test accuracy even at elevated dropout rates. In contrast, baseline methods show pronounced performance degradation, with noticeable declines in both test accuracy and convergence stability as dropout rates increase. These results underscore the robustness of FRAL-CSE in handling dynamic participation and dropouts, making it well-suited for real-world decentralized financial environments.

\section{Conclusion}
\label{con}
This work introduced a FL framework that integrates distortion risk measures to enhance robustness and interpretability in financial decision-making. By capturing tail risks, our approach ensures resilience under extreme market conditions while maintaining privacy and regulatory compliance. A key innovation is the sensitivity-aware FL training procedure, where local clients compute differential sensitivity measures, and the central server estimates global risk-aware sensitivity to guide efficient central acceleration. This design enables scalable and collaborative model training, allowing institutions to leverage decentralized knowledge without sharing proprietary data. Extensive experiments validate the effectiveness of FRAL-CSE, demonstrating superior performance in scalability, robustness to data shifts, and adaptability to dynamic client participation compared to existing baselines. By bridging financial risk management with advanced machine learning techniques, this framework provides a scalable and efficient approach to decentralized financial modeling. Future research will continue to explore more efficient central acceleration method with sparsity-drive risk sensitivity approximation.

\bibliographystyle{unsrtnat}
\bibliography{references}  

\newpage
\appendix
\onecolumn

\section{Additional Experiments}

This section presents additional experimental results that further validate the effectiveness and robustness of FRAL-CSE. These experiments are structured to align with the three primary aspects of our main analysis, i.e., scalability, robustness to data shifts, and dynamic client participation.

\subsection{Extended Evaluation on Scalability}

\begin{figure}
	\centering
	\begin{tabular}{cccc} 
		\includegraphics[width = 0.45\linewidth]{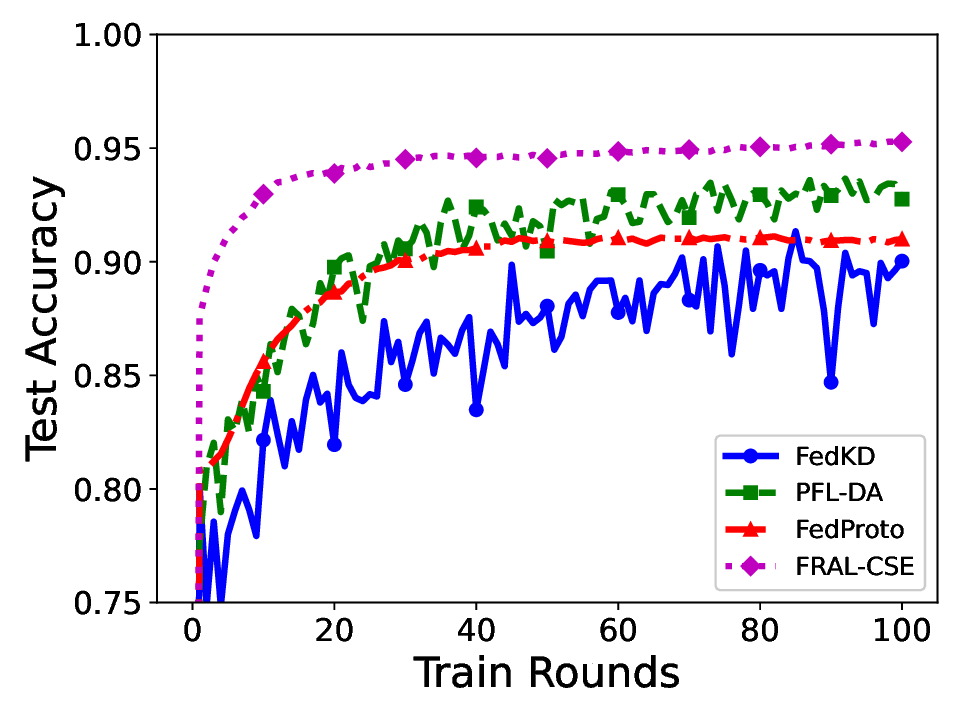}& 
		\includegraphics[width = 0.45\linewidth]{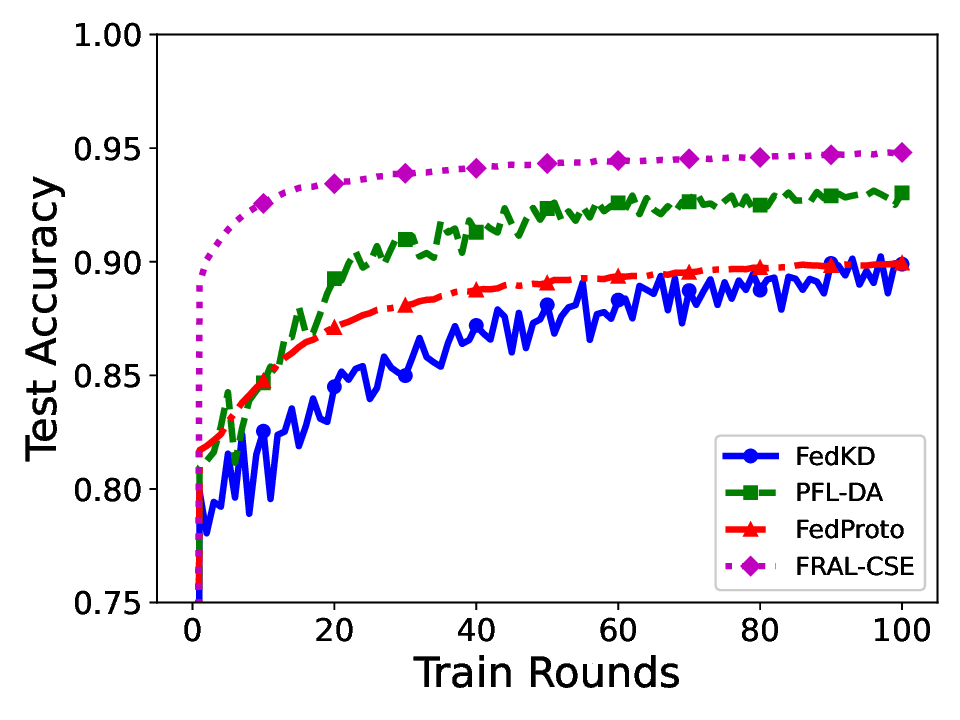} \\
		{\scriptsize (a) $30$ Clients, $30,000$ Samples Total} &
		{\scriptsize (b) $150$ Clients, $150,000$ Samples Total} 
	\end{tabular}
	\captionsetup{font={scriptsize}}
	\caption{Impact of increasing federated learning scales on test accuracy.}
	\label{fig-Scalability-acc-2}
\end{figure}

\begin{figure}
	\centering
	\begin{tabular}{cccc} 
		\includegraphics[width = 0.45\linewidth]{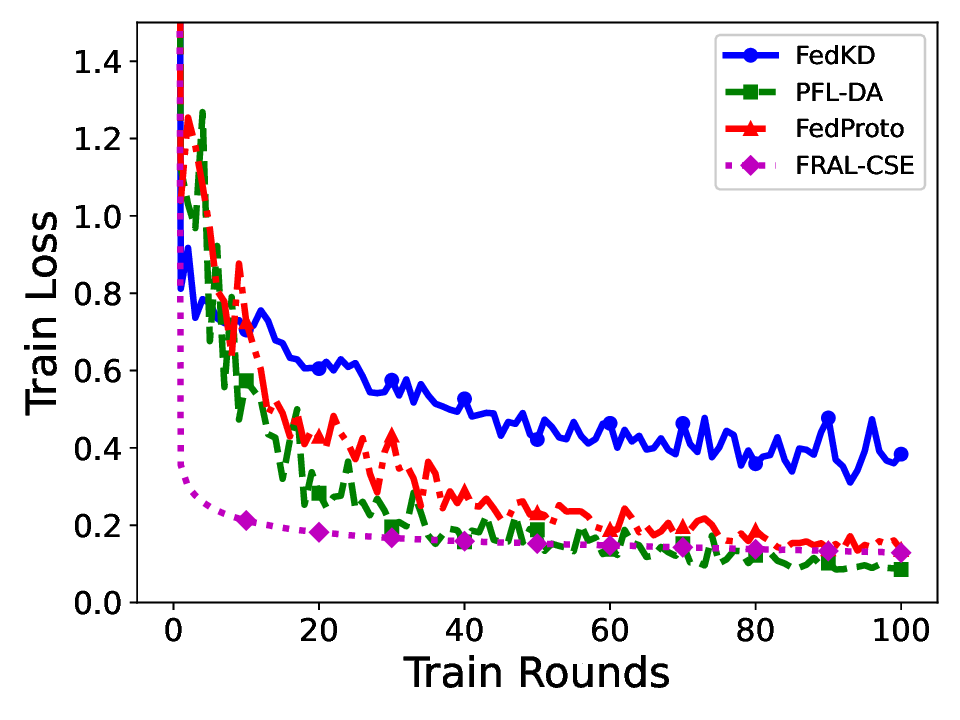}& 
		\includegraphics[width = 0.45\linewidth]{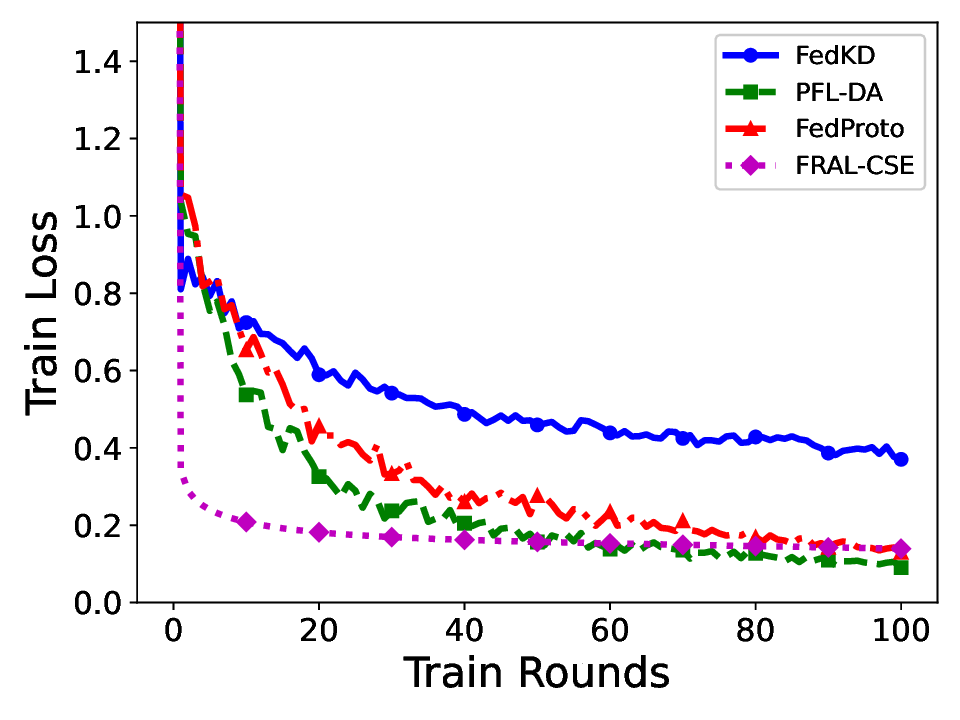} \\
		{\scriptsize (a) $30$ Clients, $30,000$ Samples Total} &
		{\scriptsize (b) $150$ Clients, $150,000$ Samples Total}
	\end{tabular}
	\captionsetup{font={scriptsize}}
	\caption{Impact of increasing federated learning scales on training loss.}
	\label{fig-Scalability-loss-2}
\end{figure}

To further evaluate the scalability of FRAL-CSE, we extend the experiments to scenarios involving diversified FL scales. Specifically, we test settings with 30 and 150 clients, as shown in Fig.~\ref{fig-Scalability-acc-2} and Fig.~\ref{fig-Scalability-loss-2}, to analyze the impact of increased collaboration on both test accuracy and training loss convergence.

Test accuracy results are shown in Fig.~\ref{fig-Scalability-acc-2}. For 30 clients, FRAL-CSE achieves significantly higher test accuracy compared to the baselines, with faster convergence and less variance during training. When the number of clients increases to $150$, FRAL-CSE maintains its advantage, demonstrating robust performance despite the increased complexity introduced by larger-scale collaboration. In contrast, baseline methods exhibit slower convergence and reduced accuracy, particularly under the higher client scale, highlighting their limitations in handling larger collaborative setups.

Training loss convergence is presented in Fig.~\ref{fig-Scalability-loss-2}. For 30 clients, FRAL-CSE demonstrates rapid convergence, achieving significantly lower training loss compared to all baseline methods. The baseline methods, particularly FedKD, exhibit noticeably slower convergence, highlighting their inefficiency in optimizing the global model under limited client participation.
As the number of clients increases to 150, FRAL-CSE continues to achieve the fastest convergence rate, effectively leveraging the additional client data to improve optimization efficiency. In contrast, the baseline methods, such as FedProto and PFL-DA, require more iterations to reach comparable training loss levels. While PFL-DA eventually converges to a slightly lower training loss than FRAL-CSE, this advantage does not translate into superior test accuracy. As shown in Fig.~\ref{fig-Scalability-acc-2}, FRAL-CSE consistently outperforms these baselines in test accuracy, underscoring its superior generalization ability and robustness to distribution shifts. This contrast highlights the limitations of the baselines, which struggle to maintain stability and adaptability in the dynamic test environment.

\subsection{Extended Evaluation on Robustness to Data Distribution Shifts}

To further evaluate the robustness of FRAL-CSE, we extend the analysis to varying client scales and train-test configurations. Specifically, we analyze performance under 10, 100, and 200 clients using two train-test splits, i.e.,  90 percent training and 10 percent testing, and 50 percent training and 50 percent testing. These configurations introduce progressively larger temporal gaps between training and testing data, simulating increasingly challenging generalization scenarios.

\begin{figure}
	\centering
	\begin{tabular}{cccc} 
		\includegraphics[width = 0.45\linewidth]{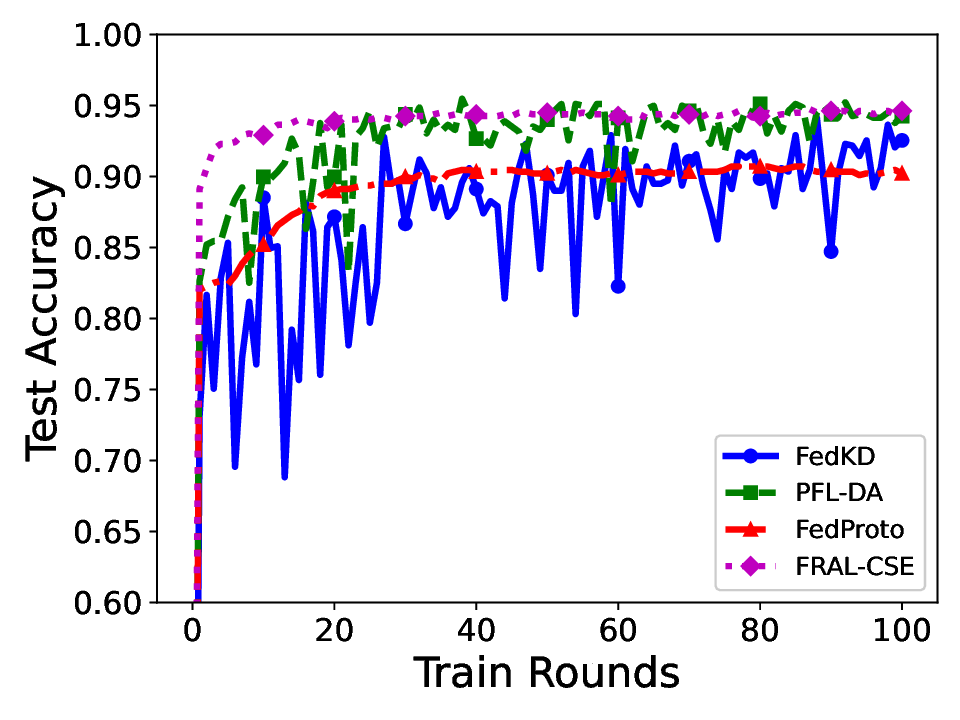} & 
		\includegraphics[width = 0.45\linewidth]{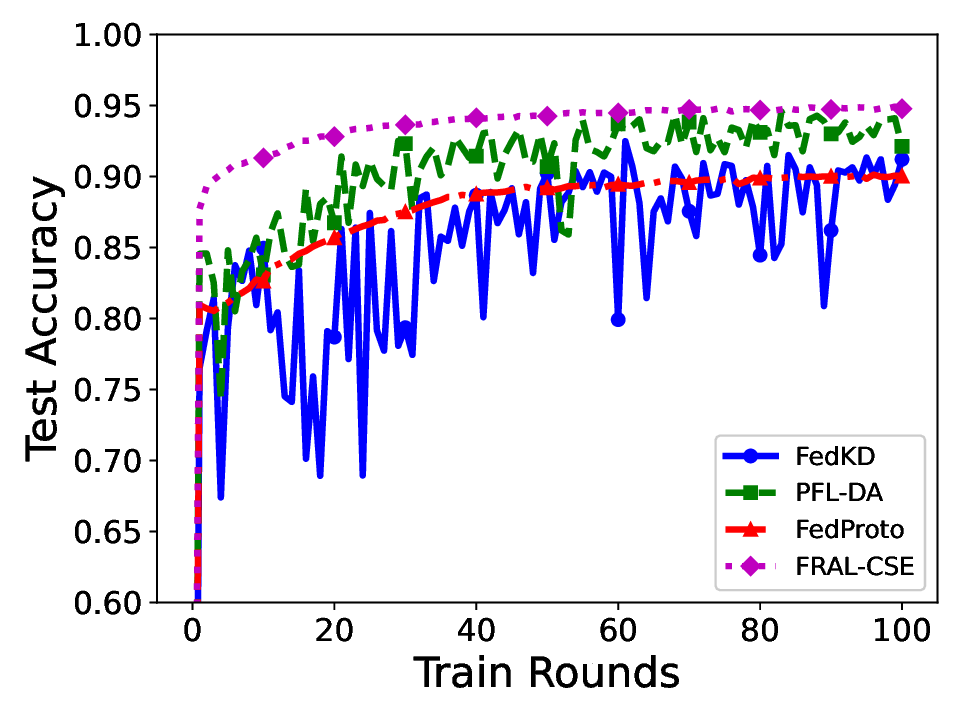} \\
		{\scriptsize (a) Training: 90\%, Testing: 10\%} &
		{\scriptsize (b) Training: 50\%, Testing: 50\%} 
	\end{tabular}
	\captionsetup{font={scriptsize}}
	\caption{Impact of different train-test splits on test accuracy with 10 clients.}
	\label{fig-Robustness-acc-10000}
\end{figure}

For the smallest scale of 10 clients, as shown in Fig.~\ref{fig-Robustness-acc-10000}, FRAL-CSE maintains its strong generalization ability despite the limited participation. In both the 90 percent training and 10 percent testing split and the 50 percent training and 50 percent testing split, FRAL-CSE achieves superior accuracy and faster convergence compared to the baseline methods. The baselines, particularly FedKD, exhibit greater instability and slower convergence, highlighting their challenges in adapting to larger data shifts.

\begin{figure}
	\centering
	\begin{tabular}{cccc} 
		\includegraphics[width = 0.45\linewidth]{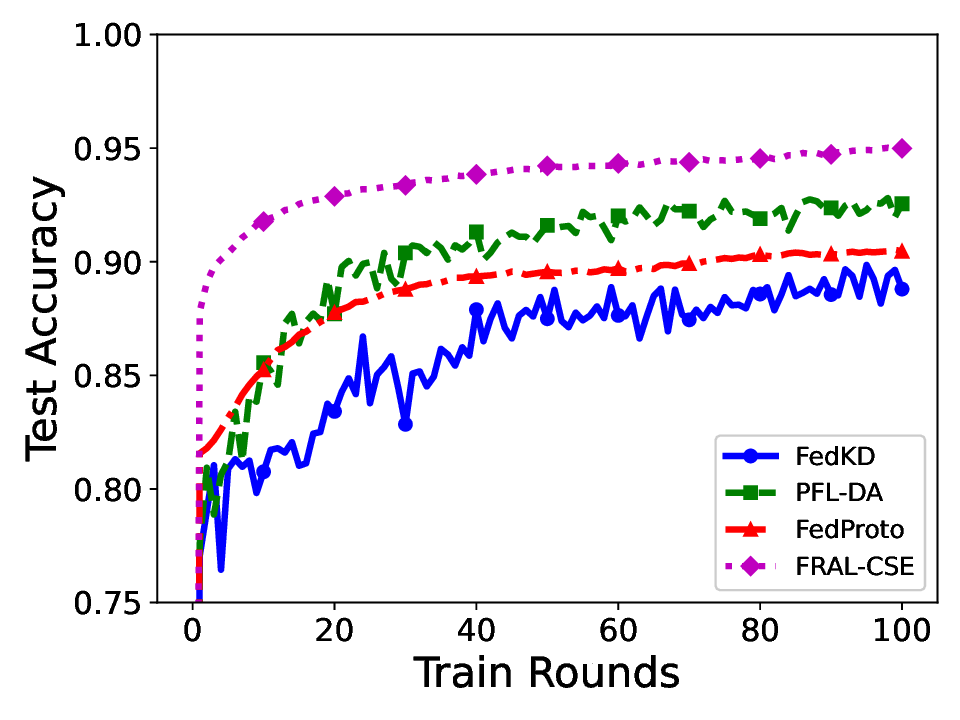} & 
		\includegraphics[width = 0.45\linewidth]{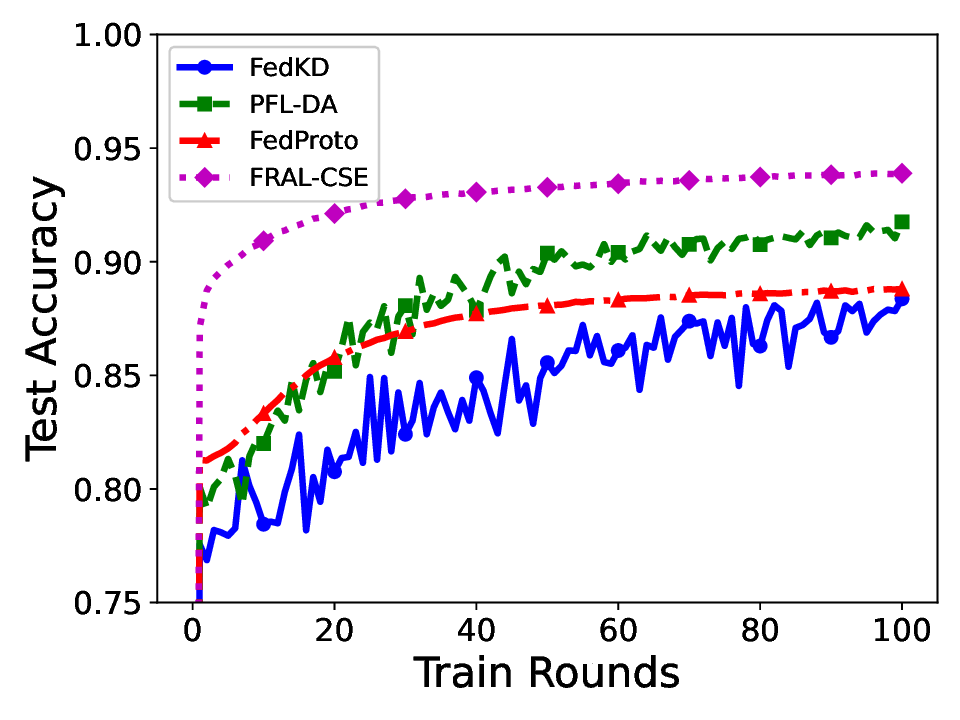} \\
		{\scriptsize (a) Training: 90\%, Testing: 10\%} &
		{\scriptsize (b) Training: 50\%, Testing: 50\%} 
	\end{tabular}
	\captionsetup{font={scriptsize}}
	\caption{Impact of different train-test splits on test accuracy with 100 clients.}
	\label{fig-Robustness-acc-100000}
\end{figure}
At a larger scale of 100 clients, as shown in Fig.~\ref{fig-Robustness-acc-100000}, FRAL-CSE continues to outperform the baselines under both train-test splits. In the 90 percent training and 10 percent testing configuration, FRAL-CSE demonstrates smoother convergence and higher accuracy across all training rounds. For the more challenging 50 percent training and 50 percent testing split, FRAL-CSE retains its advantage, showing robust generalization despite the reduced training data and greater temporal gap. The baselines struggle to achieve comparable accuracy, with FedKD showing the most pronounced instability and PFL-DA and FedProto converging more slowly.

\begin{figure}
	\centering
	\begin{tabular}{cccc} 
		\includegraphics[width = 0.45\linewidth]{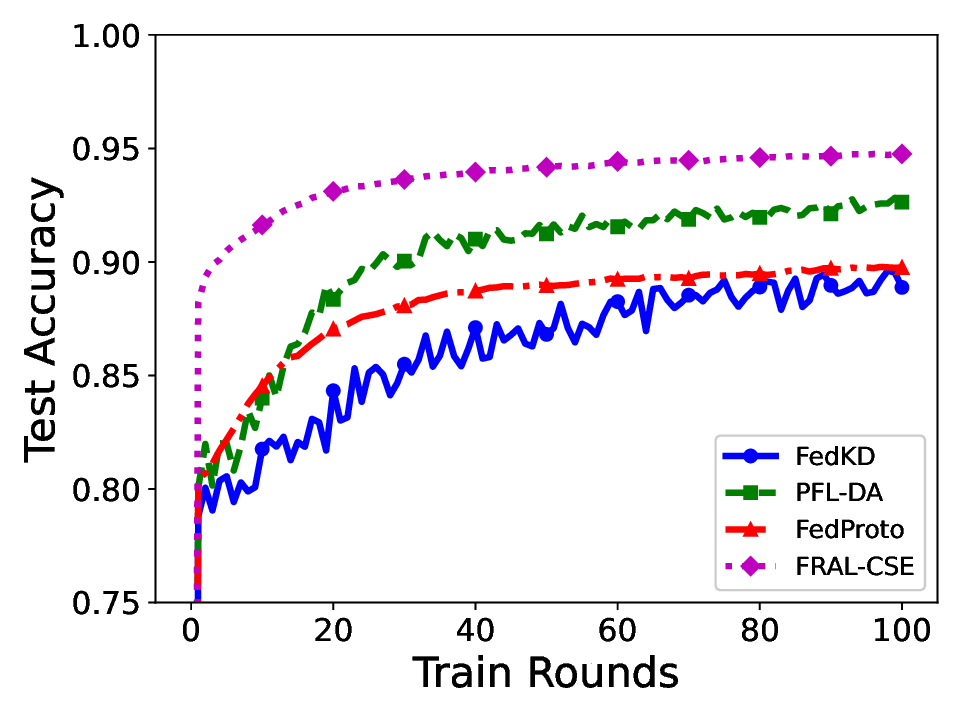} & 
		\includegraphics[width = 0.45\linewidth]{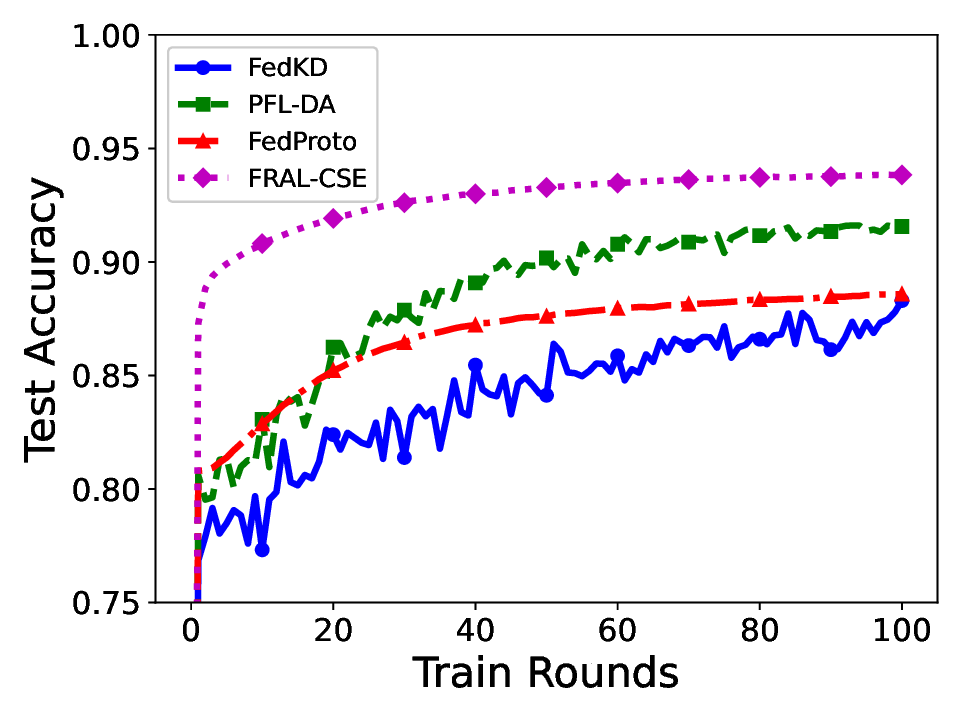} \\
		{\scriptsize (a) Training: 90\%, Testing: 10\%} &
		{\scriptsize (b) Training: 50\%, Testing: 50\%} 
	\end{tabular}
	\captionsetup{font={scriptsize}}
	\caption{Impact of different train-test splits on test accuracy with 200 clients.}
	\label{fig-Robustness-acc-200000}
\end{figure}
We further extend the analysis to 200 clients scale, as shown in Fig.~\ref{fig-Robustness-acc-200000}, to evaluate robustness in large-scale decentralized environments. FRAL-CSE maintains its superior performance across both train-test configurations, achieving faster convergence and higher accuracy compared to the baselines. In the 90 percent training and 10 percent testing split, FRAL-CSE exhibits smoother performance and greater stability, whereas the baselines, particularly FedKD, struggle to adapt to the highly distributed setting. For the 50 percent training and 50 percent testing split, FRAL-CSE demonstrates strong generalization despite the significant temporal gap and reduced training data. The baselines exhibit noticeable instability, with FedKD performing the worst, while PFL-DA and FedProto lag behind FRAL-CSE in both convergence speed and accuracy.

\subsection{Extended Evaluation on Dynamic Client Participation}

\begin{figure}
	\centering
	\begin{tabular}{cccc} 
		\includegraphics[width = 0.45\linewidth]{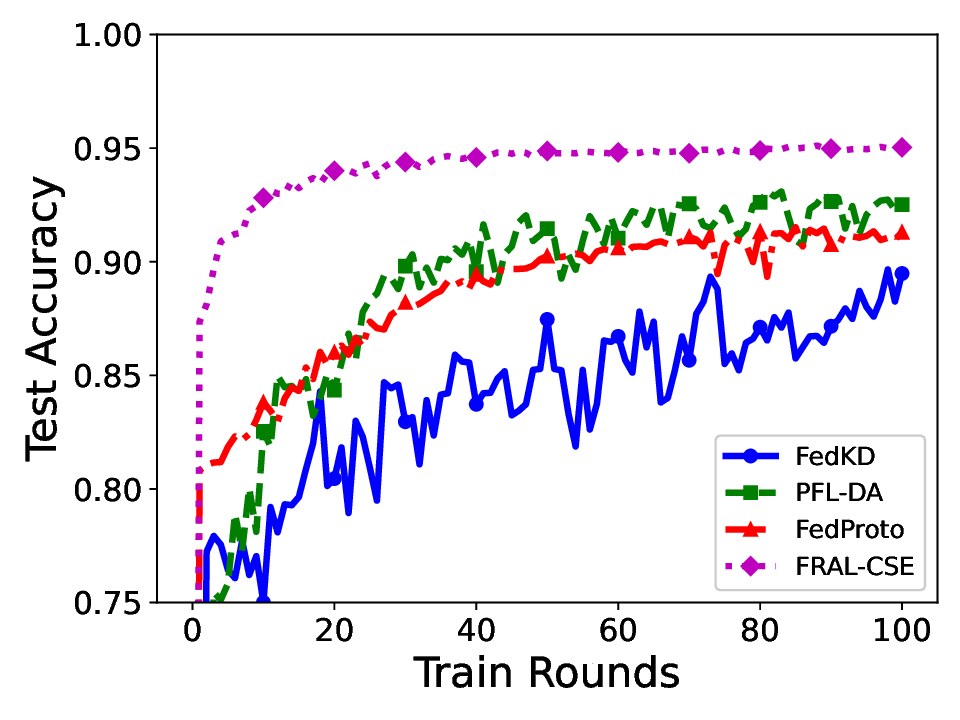} & 
		\includegraphics[width = 0.45\linewidth]{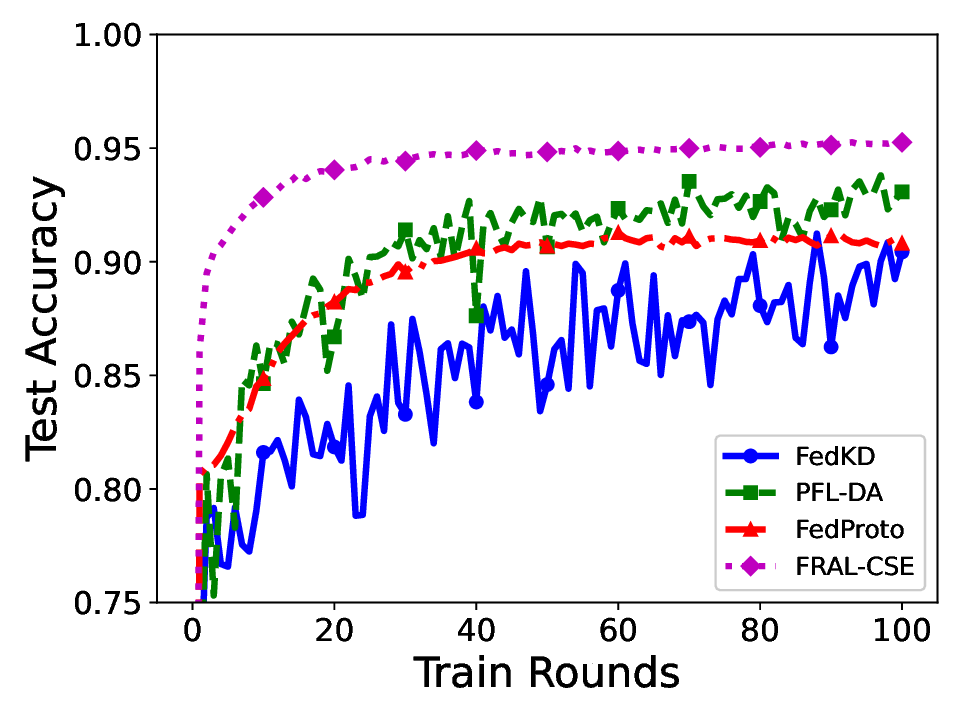} \\
		{\scriptsize (a) $20\%$ Participation Rate} &
		{\scriptsize (b) $80\%$ Participation Rate}
	\end{tabular}
	\captionsetup{font={scriptsize}}
	\caption{Impact of different participation rates on test accuracy with 30 clients.}
	\label{fig-30000-acc-jr}
\end{figure}

\begin{figure}
	\centering
	\begin{tabular}{cccc} 
		\includegraphics[width = 0.45\linewidth]{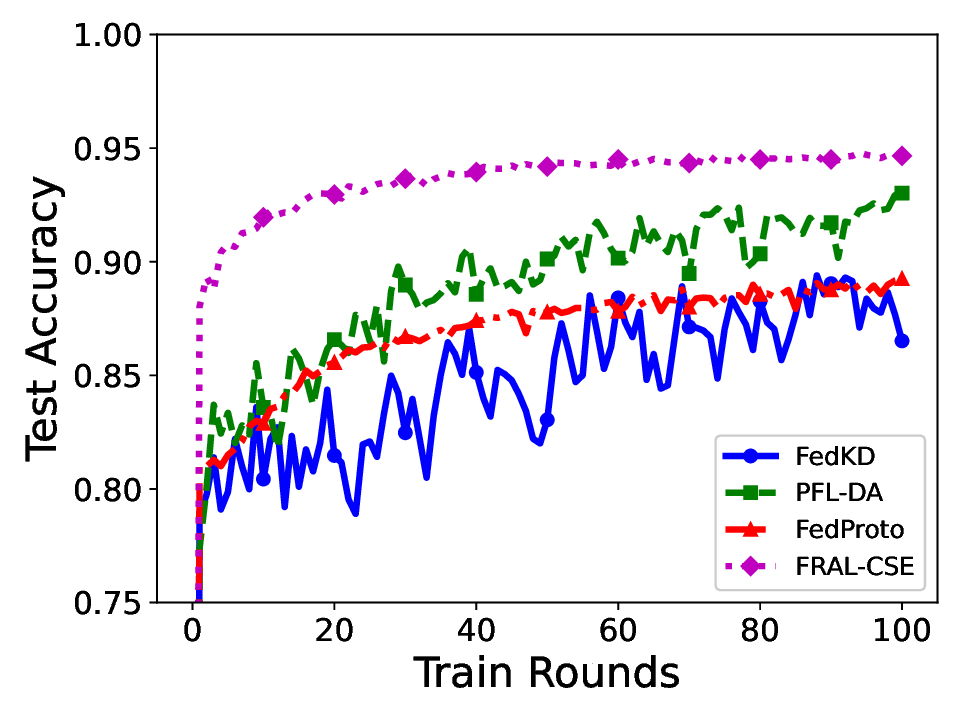} & 
		\includegraphics[width = 0.45\linewidth]{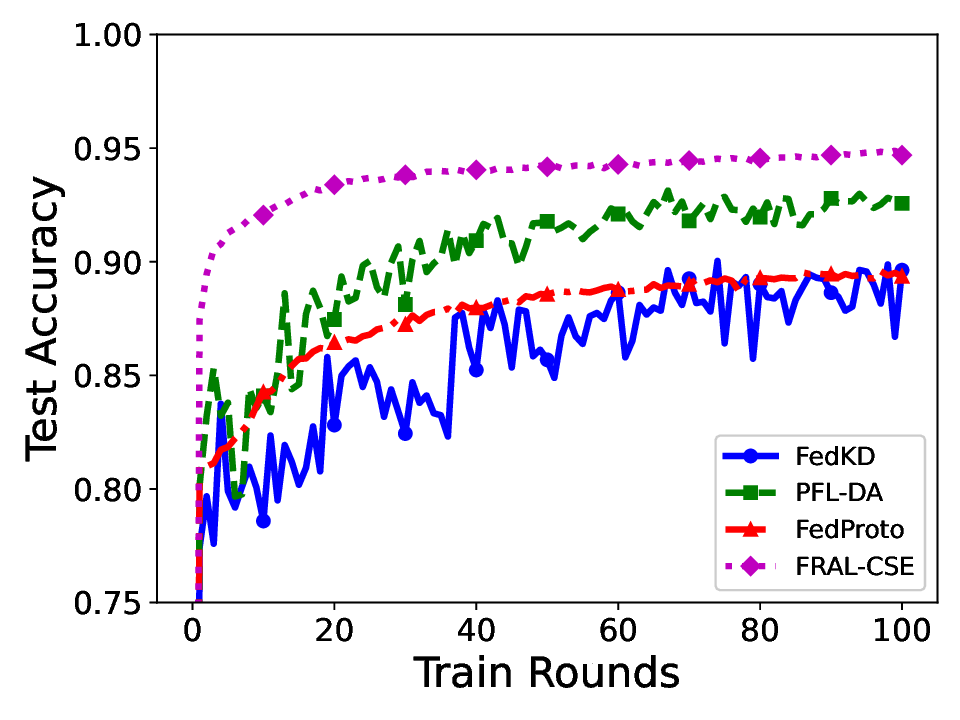} \\
		{\scriptsize (a) $20\%$ Participation Rate} &
		{\scriptsize (b) $80\%$ Participation Rate} 
	\end{tabular}
	\captionsetup{font={scriptsize}}
	\caption{Impact of different participation rates on test accuracy with 50 clients.}
	\label{fig-50000-acc-jr}
\end{figure}

\begin{figure}
	\centering
	\begin{tabular}{cccc} 
		\includegraphics[width = 0.45\linewidth]{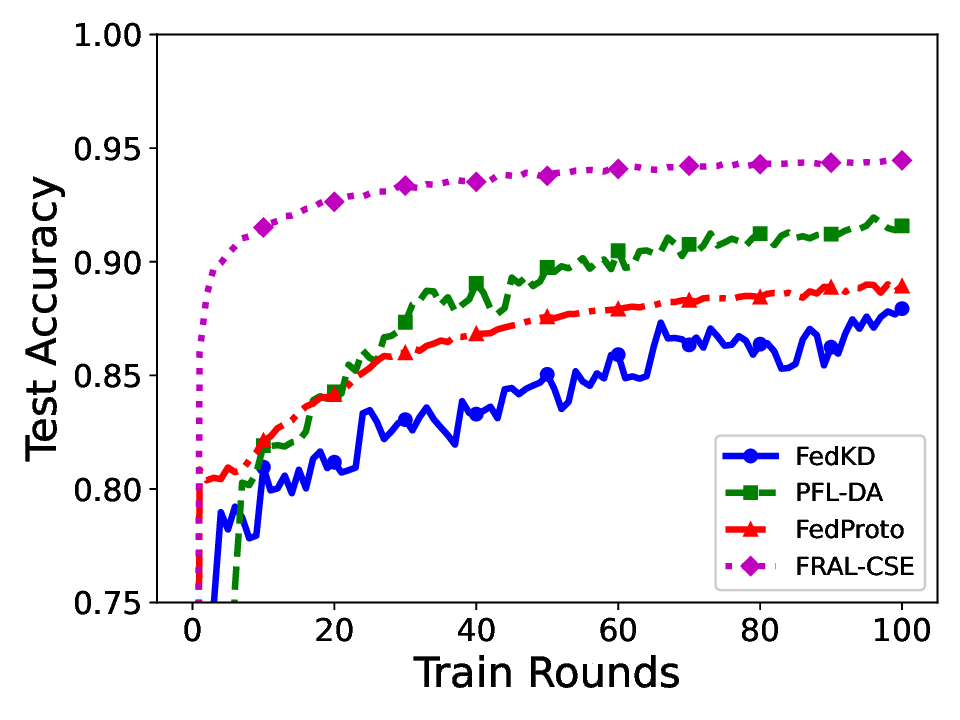} & 
		\includegraphics[width = 0.45\linewidth]{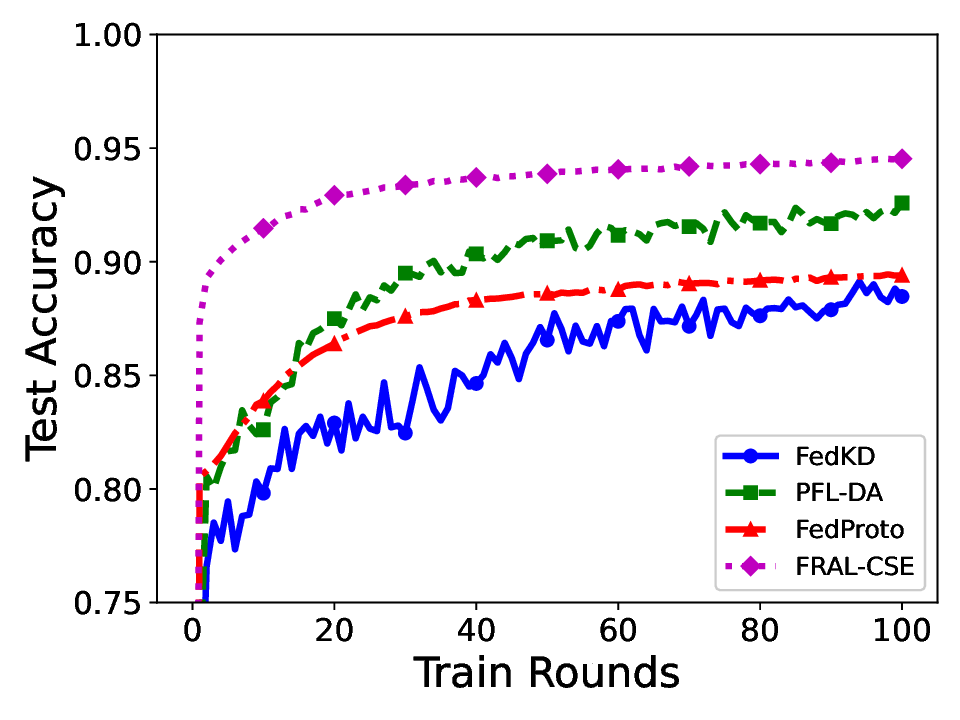} \\
		{\scriptsize (a) $20\%$ Participation Rate} &
		{\scriptsize (b) $80\%$ Participation Rate}
	\end{tabular}
	\captionsetup{font={scriptsize}}
	\caption{Performance comparison over different Participation Rate over $200$ clients.}
	\label{fig-200000-acc-jr}
\end{figure}

To further analyze the performance of FRAL-CSE, we evaluate its test accuracy under varying participation rates across three client scales, i.e., 30 clients, 50 clients, and 200 clients. For each client scale, we examine the impact of low and high client participation rates on the test accuracy. The results, shown in Fig.~\ref{fig-30000-acc-jr}, Fig.~\ref{fig-50000-acc-jr}, and Fig.~\ref{fig-200000-acc-jr}, provide insights into how FRAL-CSE maintains robust and efficient training dynamics compared to baseline methods.

For the scale of 30 clients, as shown in Fig.~\ref{fig-30000-acc-jr}, FRAL-CSE consistently outperforms the baseline methods, including FedKD, PFL-DA, and FedProto, under both participation rates. At a 20 percent participation rate, FRAL-CSE achieves faster convergence and higher test accuracy compared to the baselines, which exhibit noticeable instability and slower improvement. At the higher participation rate of 80 percent, FRAL-CSE further strengthens its advantage, maintaining smooth and efficient training dynamics while baselines such as FedKD and FedProto continue to lag behind.

As the scale increases to 50 clients, shown in Fig.~\ref{fig-50000-acc-jr}, FRAL-CSE retains its superior performance. Under the 20 percent participation rate, FRAL-CSE demonstrates strong resilience, achieving high accuracy with stable convergence, while the baselines experience significant variability and slower convergence. At the 80 percent participation rate, FRAL-CSE leverages the increased client contributions to further improve accuracy and convergence speed, outperforming the baselines across all training rounds.

In the largest scale of 200 clients, depicted in Fig.~\ref{fig-200000-acc-jr}, FRAL-CSE continues to lead in performance. At the 20 percent participation rate, the framework effectively handles the substantial data heterogeneity, achieving higher accuracy and smoother convergence than the baselines. At the 80 percent participation rate, FRAL-CSE further demonstrates its scalability, maintaining superior accuracy and convergence stability, while the baselines, particularly FedKD, struggle with large variability and slower convergence.

\begin{figure}
	\centering
	\begin{tabular}{cccc} 
		\includegraphics[width = 0.45\linewidth]{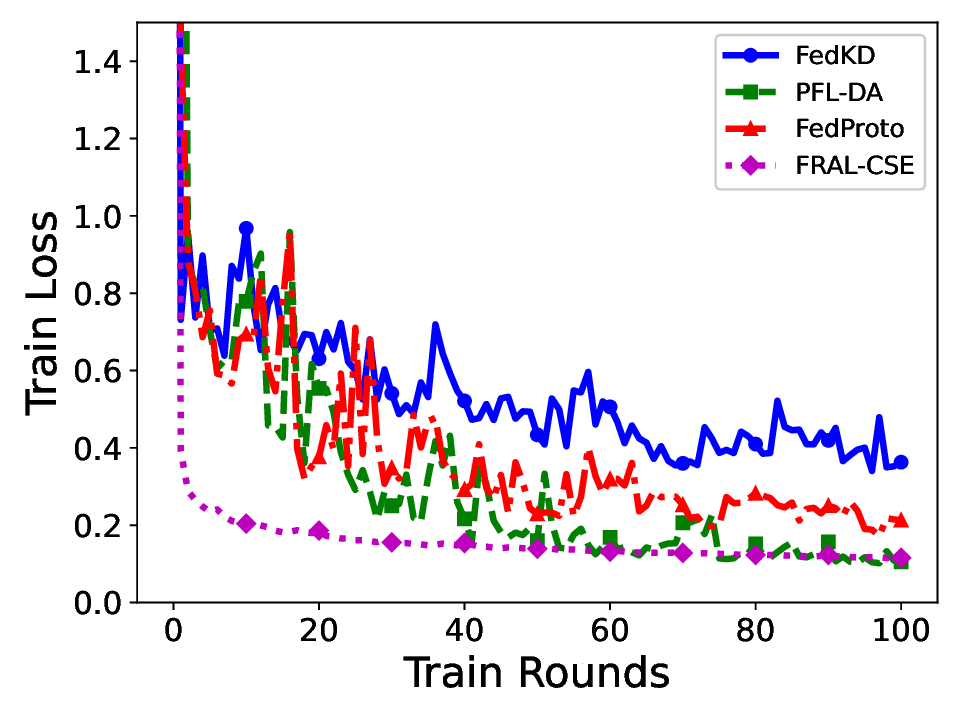} & 
		\includegraphics[width = 0.45\linewidth]{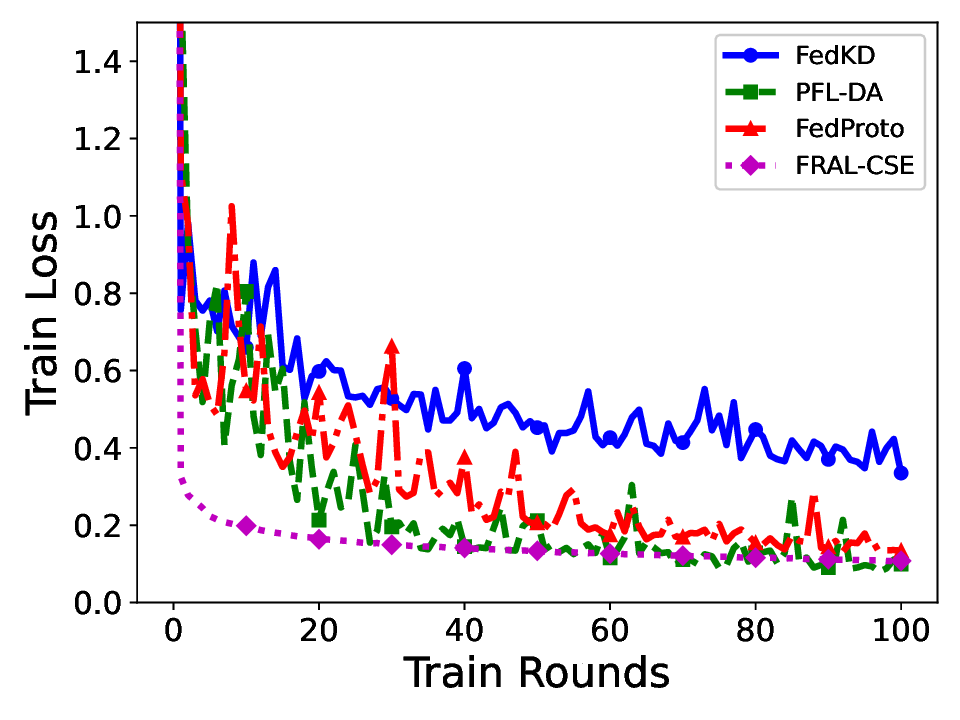} \\
		{\scriptsize (a) $20\%$ Participation Rate} &
		{\scriptsize (b) $80\%$ Participation Rate}
	\end{tabular}
	\captionsetup{font={scriptsize}}
	\caption{Impact of different participation rates on training loss with 10 clients.}
	\label{fig-10000-loss-jr}
\end{figure}

\begin{figure}
	\centering
	\begin{tabular}{cccc} 
		\includegraphics[width = 0.45\linewidth]{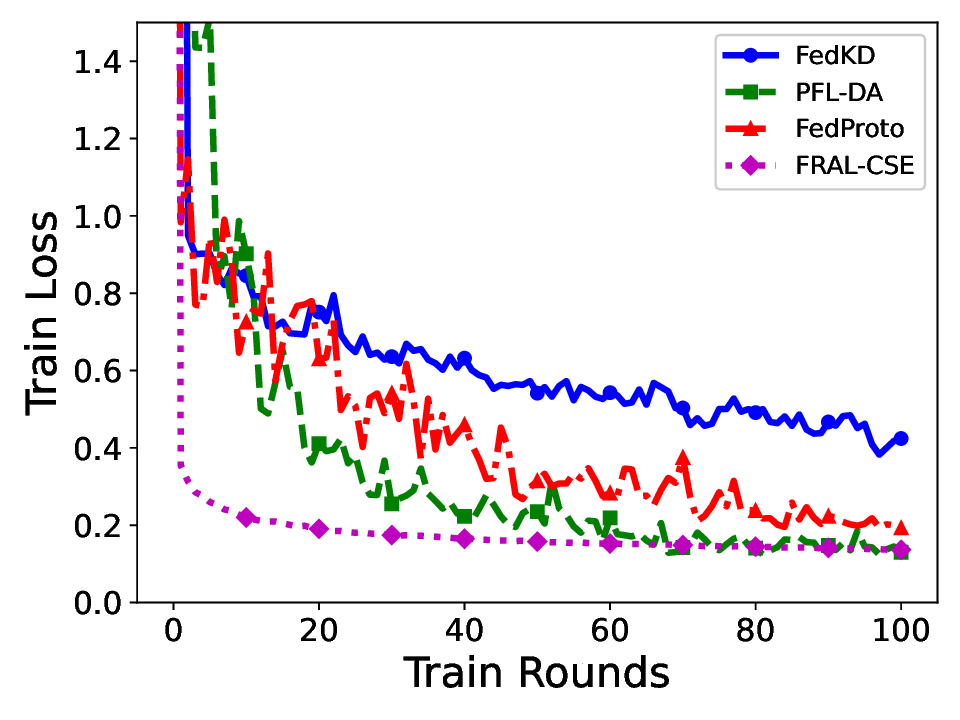} & 
		\includegraphics[width = 0.45\linewidth]{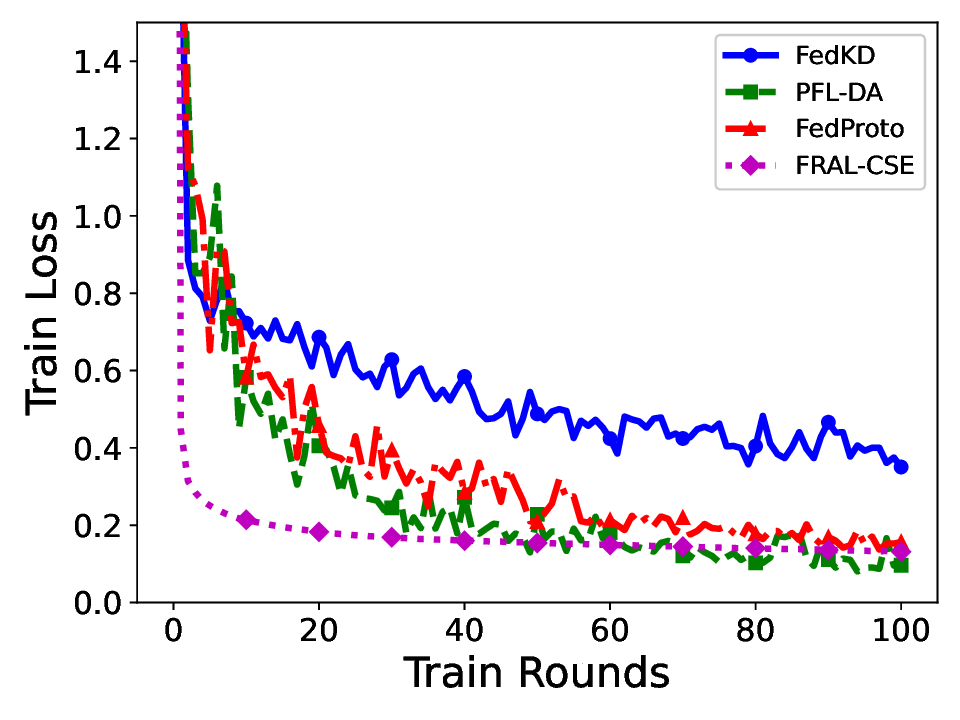} \\
		{\scriptsize (a) $20\%$ Participation Rate} &
		{\scriptsize (b) $80\%$ Participation Rate} 
	\end{tabular}
	\captionsetup{font={scriptsize}}
	\caption{Impact of different participation rates on training loss with 30 clients.}
	\label{fig-30000-loss-jr}
\end{figure}

\begin{figure}
	\centering
	\begin{tabular}{cccc} 
		\includegraphics[width = 0.45\linewidth]{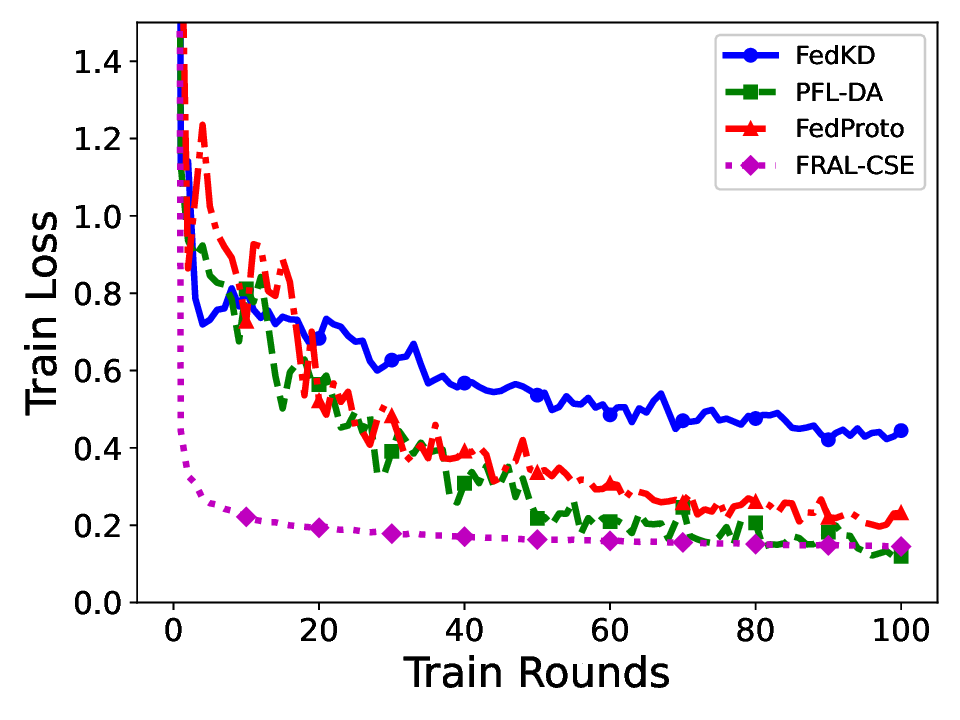} & 
		\includegraphics[width = 0.45\linewidth]{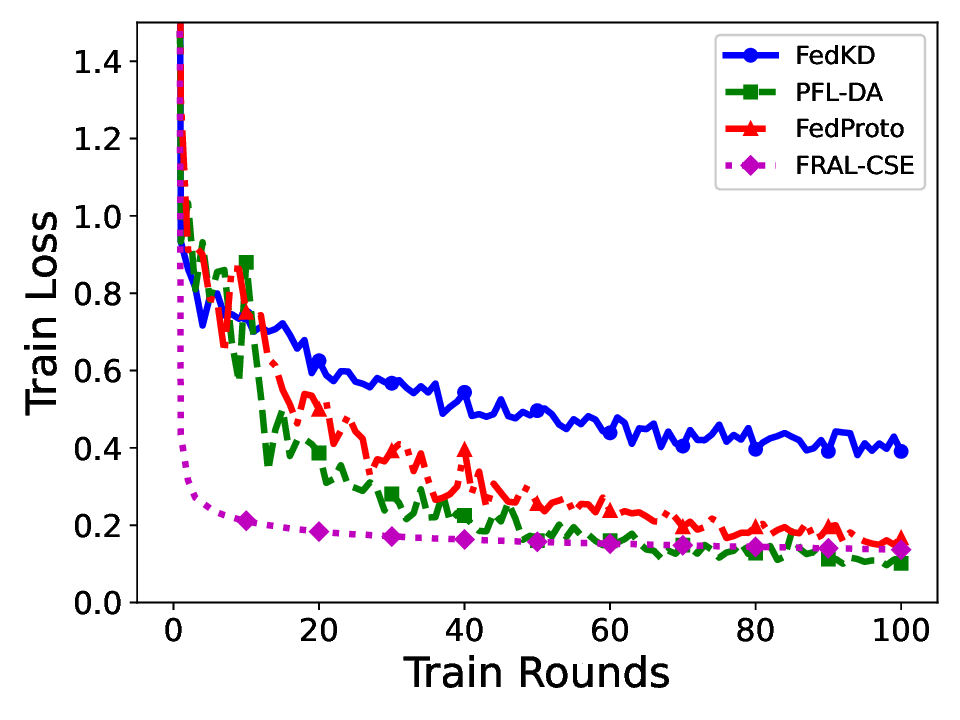} \\
		{\scriptsize (a) $20\%$ Participation Rate} &
		{\scriptsize (b) $80\%$ Participation Rate} 
	\end{tabular}
	\captionsetup{font={scriptsize}}
	\caption{Impact of different participation rates on training loss with 50 clients.}
	\label{fig-50000-loss-jr}
\end{figure}

\begin{figure}
	\centering
	\begin{tabular}{cccc} 
		\includegraphics[width = 0.45\linewidth]{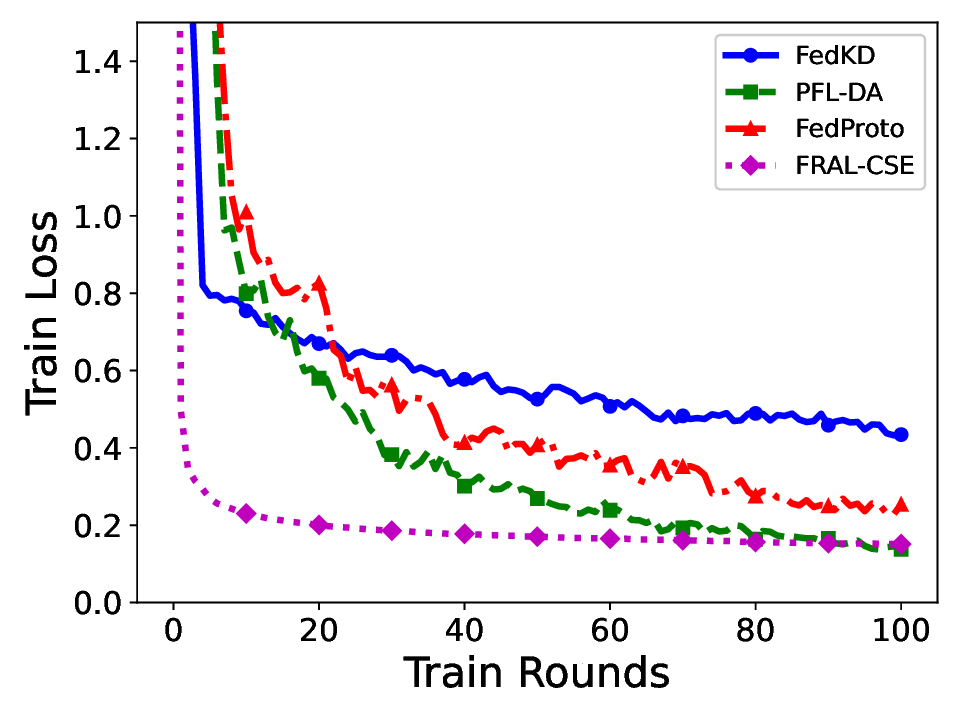} & 
		\includegraphics[width = 0.45\linewidth]{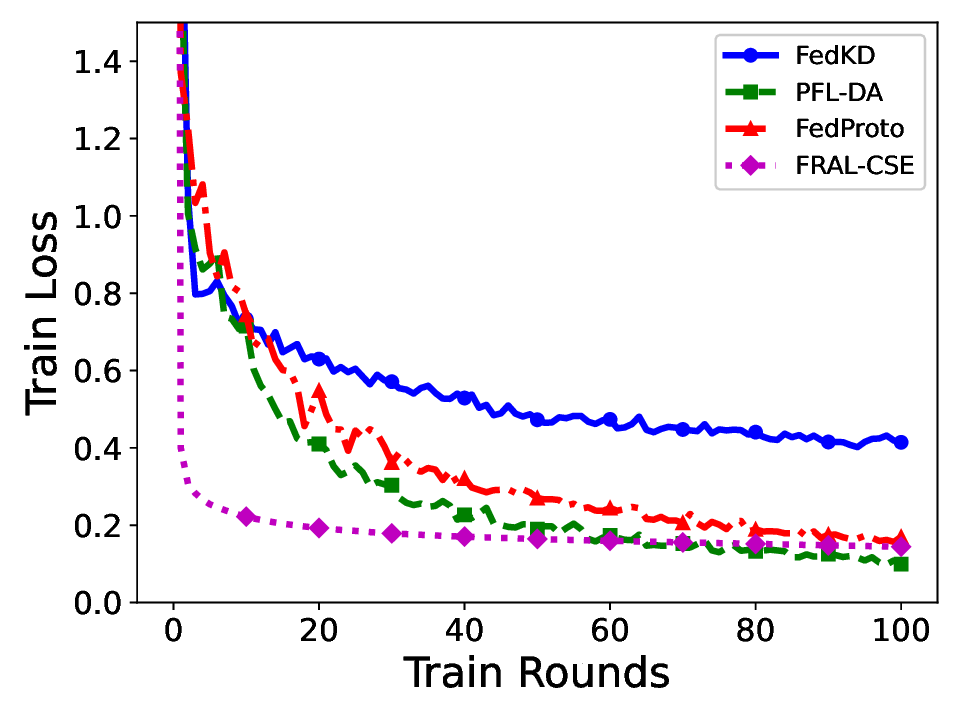} \\
		{\scriptsize (a) $20\%$ Participation Rate} &
		{\scriptsize (b) $80\%$ Participation Rate}
	\end{tabular}
	\captionsetup{font={scriptsize}}
	\caption{Impact of different participation rates on training loss with 200 clients.}
	\label{fig-200000-loss-jr}
\end{figure}

To complement the test accuracy analysis, we further assess the training loss convergence of FRAL-CSE and baseline methods across varying participation rates and client scales. The results for 10, 30, 50, and 200 clients are shown in Fig.~\ref{fig-10000-loss-jr}, Fig.~\ref{fig-30000-loss-jr}, Fig.~\ref{fig-50000-loss-jr}, and Fig.~\ref{fig-200000-loss-jr}, respectively.

At the 10-client scale as shown in Fig.~\ref{fig-10000-loss-jr}, FRAL-CSE exhibits significantly lower training loss than all baselines under both participation rates. At 20 percent participation, it achieves rapid and stable convergence, while FedKD and FedProto show slower optimization and higher variability. As participation increases to 80 percent, FRAL-CSE maintains its advantage, continuing to converge faster and more smoothly than the baselines, which still exhibit fluctuations. 
For the 30-client scale as shown in Fig.~\ref{fig-30000-loss-jr}, FRAL-CSE remains the most efficient, achieving the lowest training loss and the most stable convergence. At 20 percent participation, it demonstrates robust optimization, while FedKD experiences significant instability. At 80 percent participation, FRAL-CSE further improves, reinforcing its efficiency and resilience compared to the baselines.
At the 50-client scale illustrated in Fig.~\ref{fig-50000-loss-jr}, FRAL-CSE continues to outperform all baselines. At 20 percent participation, it achieves the lowest training loss across all training rounds, while FedKD and FedProto suffer from higher variability and slower convergence. With 80 percent participation, FRAL-CSE further stabilizes, ensuring an efficient optimization process with minimal fluctuations, whereas the baselines remain less stable.
At the largest scale of 200 clients as shown in Fig.~\ref{fig-200000-loss-jr}, FRAL-CSE consistently outperforms the baselines across both participation rates. Under 20 percent participation, it effectively handles increased data heterogeneity, achieving significantly lower training loss and faster convergence than the baselines. At 80 percent participation, it maintains robust performance with stable optimization, while FedKD struggles with noticeable slower convergence.

\subsection{Evaluation Under Dynamic Client Drop-Out Scenarios}

\begin{figure}
	\centering
	\begin{tabular}{cccc} 
		\includegraphics[width = 0.45\linewidth]{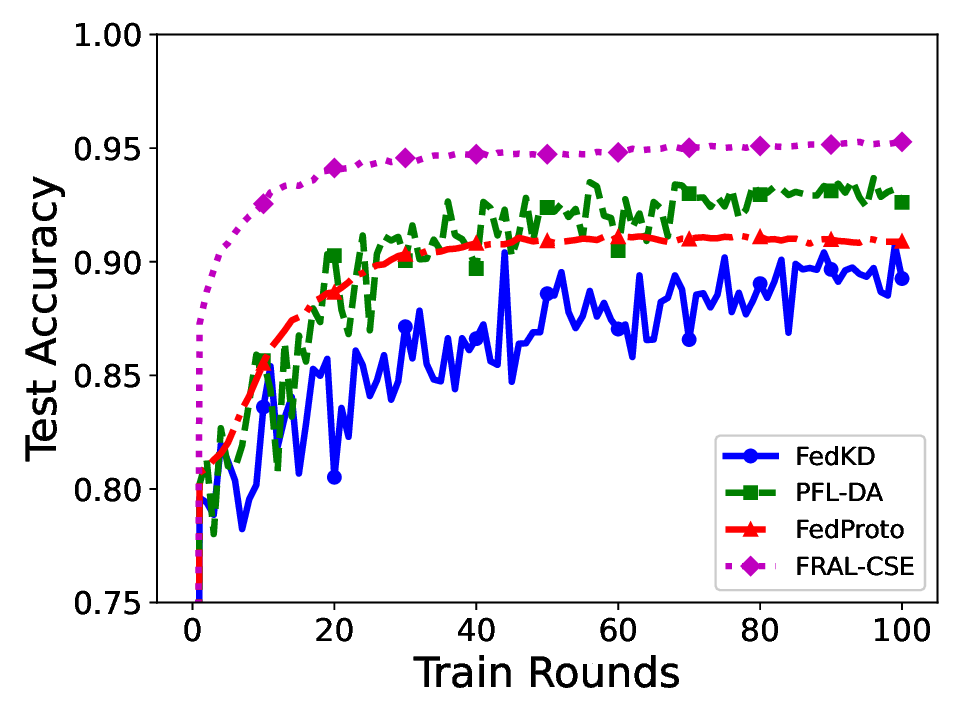} & 
		\includegraphics[width = 0.45\linewidth]{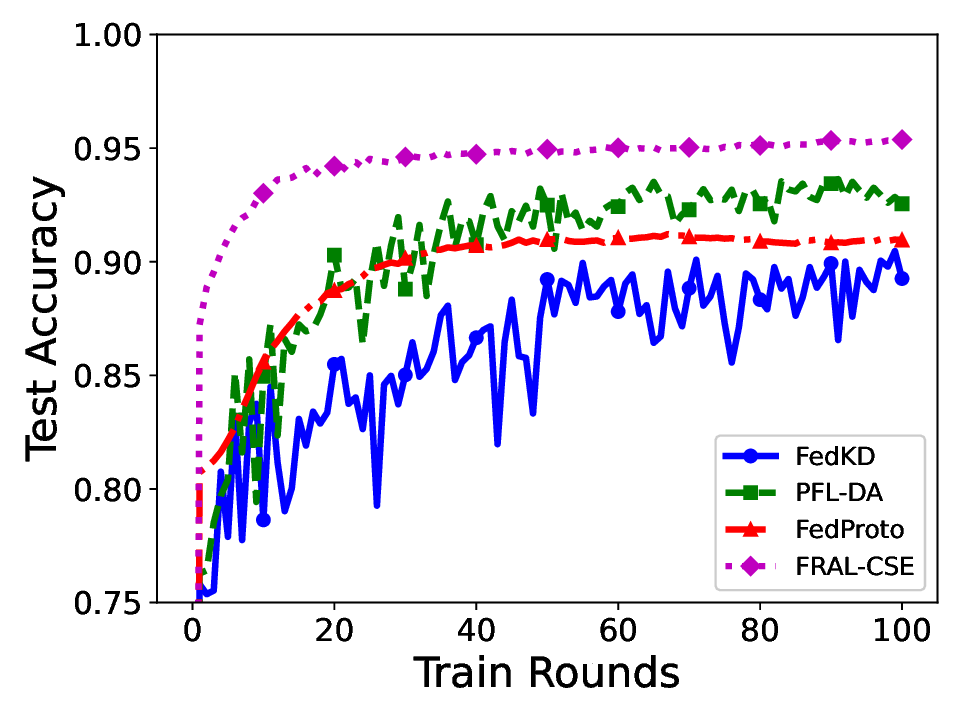} \\
		{\scriptsize (a) $10\%$ Dynamic Client Dropout} &
		{\scriptsize (b) $40\%$ Dynamic Client Dropout} 
	\end{tabular}
	\captionsetup{font={scriptsize}}
	\caption{Impact of dynamic client dropout on test accuracy with 30 clients.}
	\label{fig-drop-out-acc-30000}
\end{figure}

\begin{figure}
	\centering
	\begin{tabular}{cccc} 
		\includegraphics[width = 0.45\linewidth]{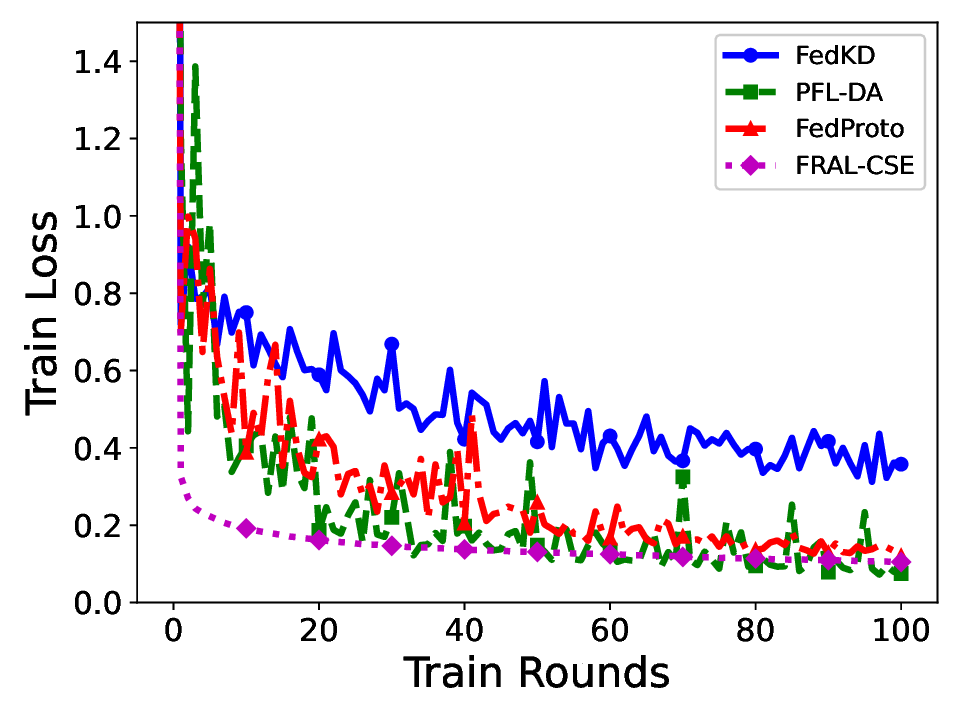} & 
		\includegraphics[width = 0.45\linewidth]{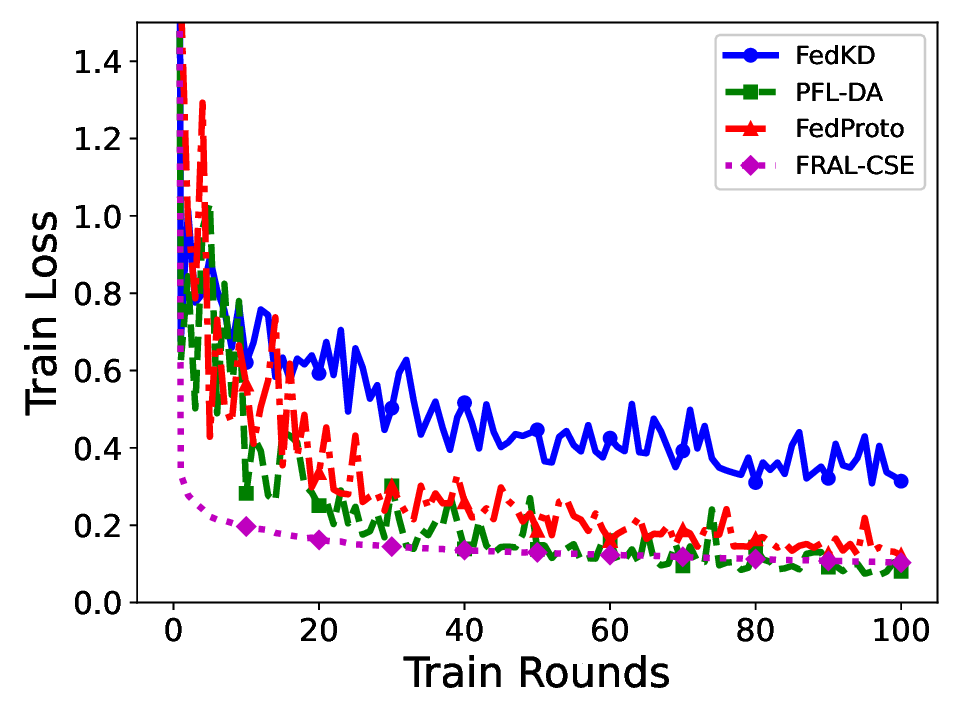} \\
		{\scriptsize (a) $10\%$ Dynamic Client Dropout} &
		{\scriptsize (b) $40\%$ Dynamic Client Dropout}
	\end{tabular}
	\captionsetup{font={scriptsize}}
	\caption{Impact of dynamic client dropout on training loss with 10 clients.}
	\label{fig-drop-out-loss-10000}
\end{figure}

\begin{figure}
	\centering
	\begin{tabular}{cccc} 
		\includegraphics[width = 0.45\linewidth]{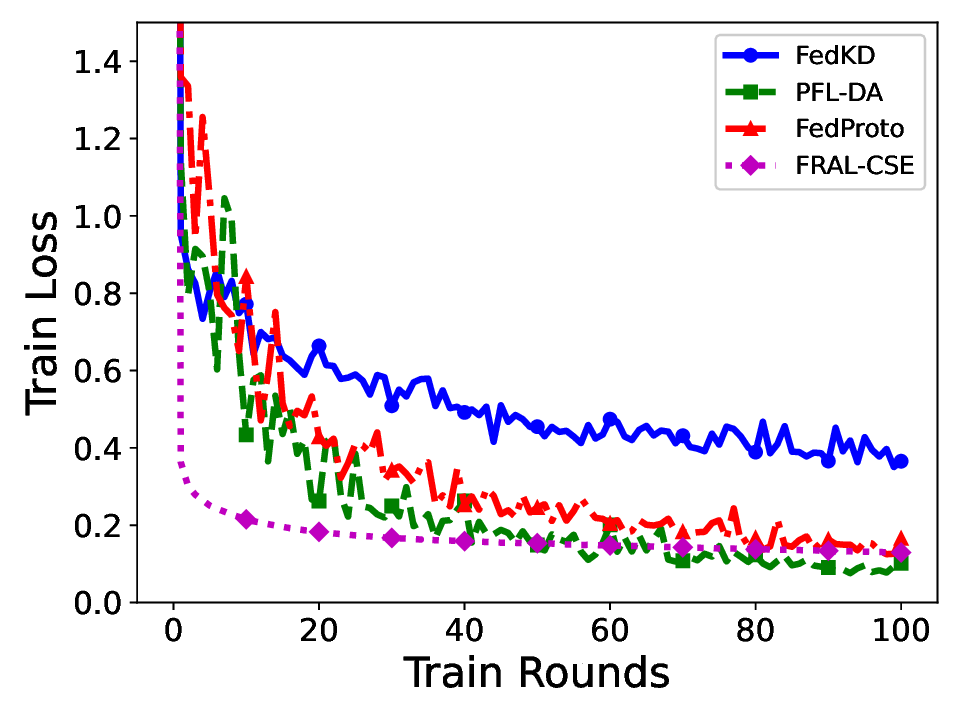} & 
		\includegraphics[width = 0.45\linewidth]{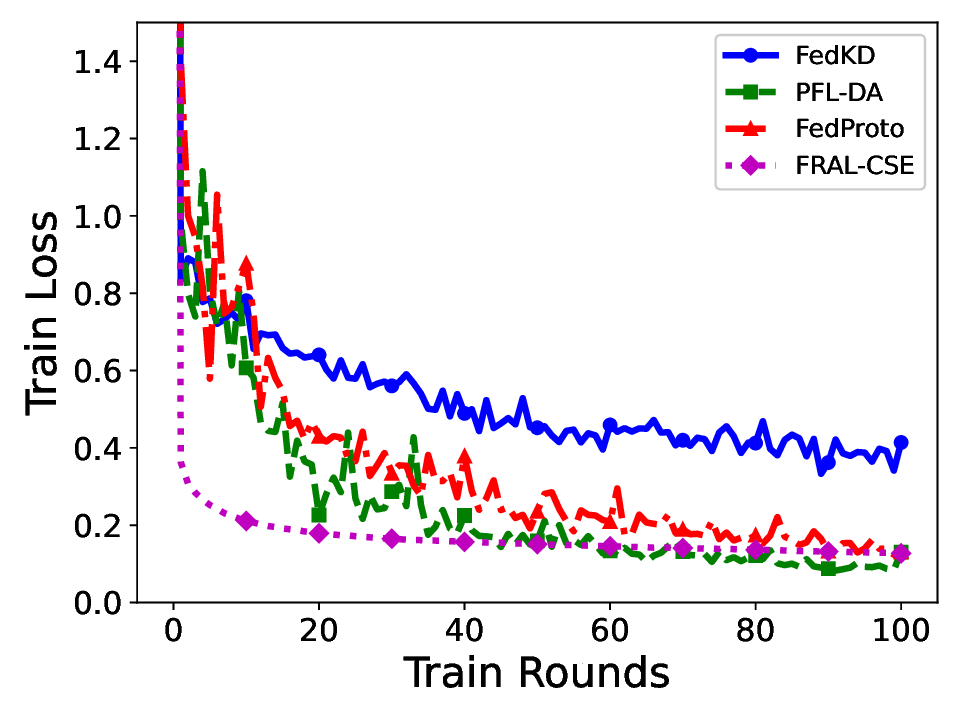} \\
		{\scriptsize (a) $10\%$ Dynamic Client Dropout} &
		{\scriptsize (b) $40\%$ Dynamic Client Dropout} 
	\end{tabular}
	\captionsetup{font={scriptsize}}
	\caption{Impact of dynamic client dropout on training loss with 30 clients.}
	\label{fig-drop-out-loss-30000}
\end{figure}

\begin{figure}
	\centering
	\begin{tabular}{cccc} 
		\includegraphics[width = 0.45\linewidth]{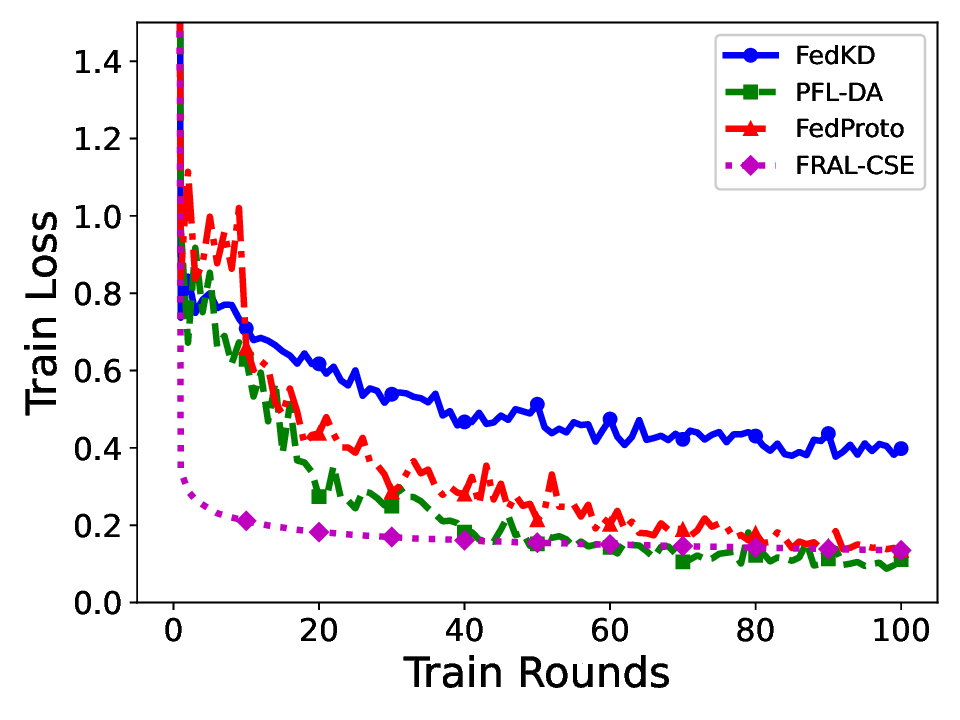} & 
		\includegraphics[width = 0.45\linewidth]{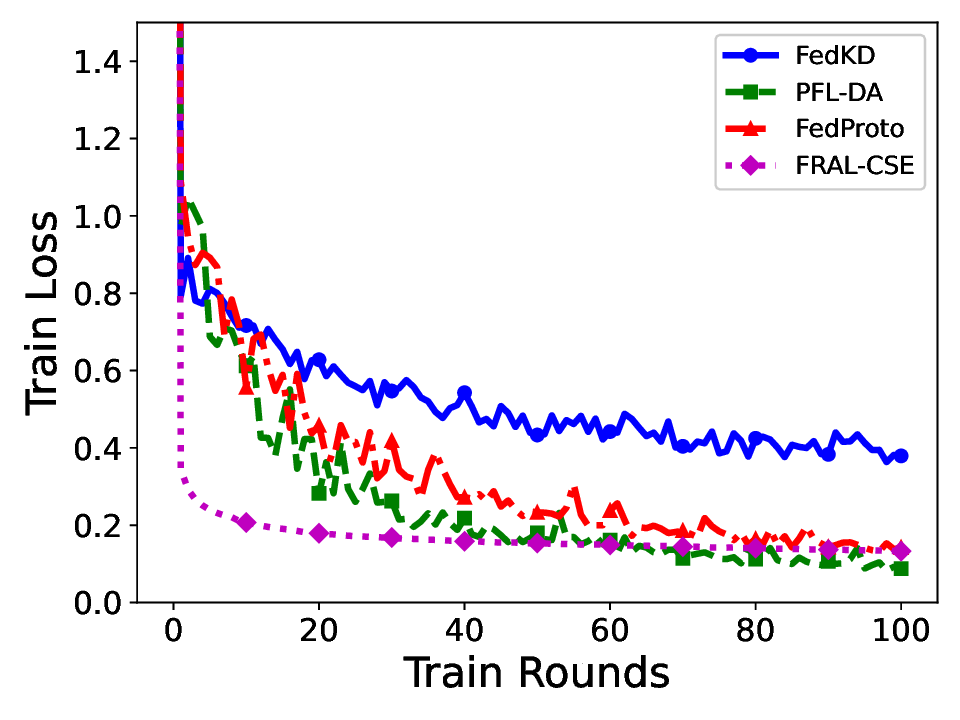} \\
		{\scriptsize (a) $10\%$ Dynamic Client Dropout} &
		{\scriptsize (b) $40\%$ Dynamic Client Dropout} 
	\end{tabular}
	\captionsetup{font={scriptsize}}
	\caption{Impact of dynamic client dropout on training loss with 50 clients.}
	\label{fig-drop-out-loss-50000}
\end{figure}

\begin{figure}
	\centering
	\begin{tabular}{cccc} 
		\includegraphics[width = 0.45\linewidth]{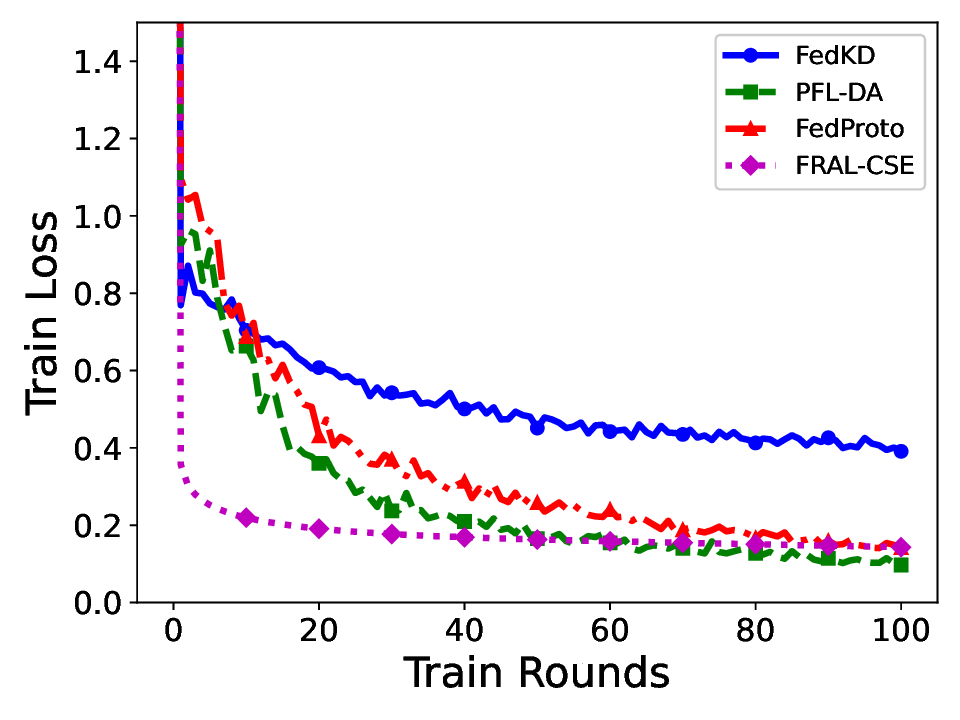} & 
		\includegraphics[width = 0.45\linewidth]{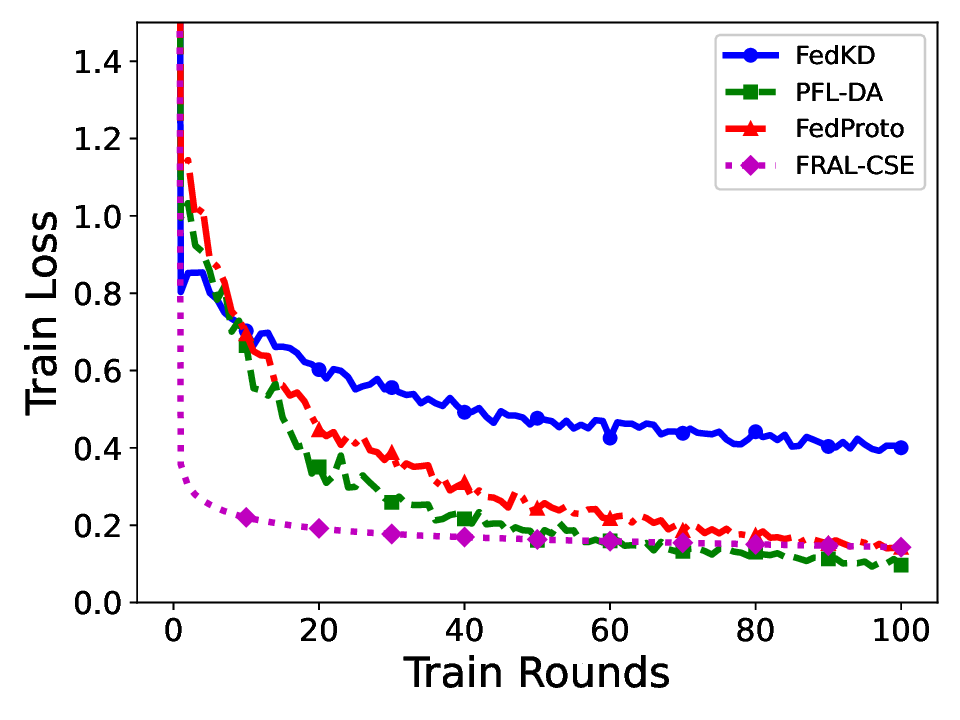} \\
		{\scriptsize (a) $10\%$ Dynamic Client Dropout} &
		{\scriptsize (b) $40\%$ Dynamic Client Dropout}  
	\end{tabular}
	\captionsetup{font={scriptsize}}
	\caption{Impact of dynamic client dropout on training loss with 200 clients.}
	\label{fig-drop-out-loss-200000}
\end{figure}

To evaluate the robustness of FRAL-CSE under dynamic client drop-out conditions, we analyze both test accuracy and training loss across varying client scales and drop-out rates. Specifically, we consider drop-out rates of 10 percent and 40 percent for 10, 30, 50, and 200 clients. The results, presented in Fig.~\ref{fig-drop-out-acc-30000} for test accuracy and Fig.~\ref{fig-drop-out-loss-10000}, Fig.~\ref{fig-drop-out-loss-30000}, Fig.~\ref{fig-drop-out-loss-50000}, and Fig.~\ref{fig-drop-out-loss-200000} for training loss, provide a comprehensive understanding of FRAL-CSE's performance compared to baseline methods.

For test accuracy with 30 clients, as shown in Fig.~\ref{fig-drop-out-acc-30000}, FRAL-CSE demonstrates strong generalization and stability across both drop-out rates. At a 10 percent drop-out rate, FRAL-CSE achieves superior accuracy with faster convergence and smoother performance compared to baseline methods, such as FedKD and FedProto, which exhibit slower convergence and higher variability. Even at a 40 percent drop-out rate, FRAL-CSE retains its robustness, maintaining high test accuracy and stable convergence, while the baselines struggle to adapt to the increased drop-out dynamics, showing significant variability and slower convergence.

For training loss with 10 clients, as shown in Fig.~\ref{fig-drop-out-loss-10000}, FRAL-CSE achieves rapid convergence with the lowest training loss across all training rounds under the 10 percent drop-out rate. In contrast, baseline methods, particularly FedKD and FedProto, exhibit greater variability and require more iterations to reach comparable loss levels. When the drop-out rate increases to 40 percent, FRAL-CSE maintains stable convergence, while the baselines struggle with instability and slower optimization.

For training loss with 30 clients, depicted in Fig.~\ref{fig-drop-out-loss-30000}, FRAL-CSE consistently outperforms the baselines across both drop-out rates, achieving faster convergence and lower training loss. At the 10 percent drop-out rate, FRAL-CSE demonstrates smooth and efficient optimization, whereas the baselines exhibit fluctuating loss curves and slower adaptation. When the drop-out rate increases to 40 percent, FRAL-CSE continues to achieve stable convergence, while FedProto and FedKD require additional training iterations to compensate for the increased client variability.

The results for 50 clients, shown in Fig.~\ref{fig-drop-out-loss-50000}, further emphasize FRAL-CSE's robustness under dynamic client participation. At the 10 percent drop-out rate, FRAL-CSE rapidly converges with the lowest training loss across all training rounds. The baselines, particularly FedKD and PFL-DA, exhibit greater variability and require extended iterations to stabilize. Under the 40 percent drop-out rate, FRAL-CSE retains its efficiency, while baseline methods continue to struggle with increased instability and slower convergence.

At the largest scale of 200 clients, as depicted in Fig.~\ref{fig-drop-out-loss-200000}, FRAL-CSE maintains its strong performance in handling dynamic drop-out scenarios. Under the 10 percent drop-out rate, FRAL-CSE achieves rapid and stable convergence with significantly lower training loss compared to the baselines. When the drop-out rate increases to 40 percent, FRAL-CSE remains highly efficient, ensuring smooth optimization across all training rounds. In contrast, baseline methods such as FedKD and FedProto experience notable instability, requiring prolonged iterations to reach comparable loss levels. Although PFL-DA eventually converges to a slightly lower training loss than FRAL-CSE, this advantage does not translate into better test accuracy. FRAL-CSE consistently outperforms all baselines in generalization, underscoring its superior adaptability to real-world financial environments with dynamic participation.

\end{document}